\documentclass[10pt]{article}
\usepackage[margin=1in]{geometry}
\usepackage[T1]{fontenc}
\usepackage{amsmath,amssymb,booktabs,graphicx,array}
\usepackage[numbers,sort&compress]{natbib}
\usepackage{hyperref}
\usepackage{lmodern}
\usepackage{lineno}

\newcommand{\nimis}{\textsc{NIMIS}}
\newcommand{\figimg}[2][\linewidth]{%
  \IfFileExists{#2}{\includegraphics[width=#1]{#2}}%
  {\fbox{\parbox{0.9\linewidth}{\centering\vspace{2.5cm}missing figure file: \texttt{#2}\vspace{2.5cm}}}}}

\title{\nimis: A Structure-Agnostic, Chemically Informed Slot-Attention Framework for Demixing 1D $^1$H-NMR Mixture Spectra}
\author{Anna Marija Sumrova \and Jonathan M.\ Goodman\\[0.4em]
Yusuf Hamied Department of Chemistry, Lensfield Road, Cambridge CB2 1EW}
\date{}

\begin{document}
\maketitle

\begin{abstract}
The products of chemical reactions are usually complex mixtures that must be separated into their components before interpretable $^1$H-NMR spectra can be obtained. Direct analysis of the spectra of crude mixtures would accelerate reaction analysis and facilitate the exploration of new chemical space. We present \nimis, a structure-agnostic, attention-based deep learning framework that groups the peaks of a 1D $^1$H-NMR mixture spectrum according to the component from which they originate. J-coupling information is introduced directly into the attention mechanism of a multi-head Transformer encoder, and the resulting contextualised peak embeddings are separated into components by slot attention. \nimis{} was trained on 158,005 synthetic mixtures constructed from experimental spectra in the NMRexp database, covering molecules of up to 20 heavy atoms composed of C, H, N, O, S, F, Cl, Br and I. On held-out molecules it reaches a peak-to-component assignment accuracy of 82.28\% for noise-free input and 77.34\% with 10\% integral noise, with a mean \nimis{} score of 0.83 for correct assignments. \nimis{} outperforms a conventional Transformer and simple baselines, and its performance carries over to a scaffold-disjoint evaluation. By grouping peaks into component-level sets, \nimis{} provides a foundation for more robust mixture analysis methods, including methods capable of handling overlapping signals.
\end{abstract}

\section{Introduction}

Nuclear magnetic resonance (NMR) spectroscopy is a powerful analytical technique used for both qualitative and quantitative studies across the molecular sciences. Interpreting NMR spectra, however, requires substantial chemical expertise. Recent advances in cheminformatics, including statistical and artificial intelligence approaches, have enabled automated uncertainty estimation for structure-to-spectrum matching~\citep{r1} and even direct structure prediction from spectral data~\citep{r2,r3,r4}. Most existing approaches, however, focus on spectra of single compounds. Obtaining such spectra typically requires considerable synthetic effort to isolate pure compounds, and thus assumes a high level of knowledge about the system under study. In contrast, chemical mixtures are far more common than isolated molecules, but their spectra are considerably harder to interpret, even for experienced chemists. Severe peak overlap and other spectral complexity often lead to abandonment of the analysis or to modification of the synthetic experiment. This limitation is a substantial bottleneck in reaction analysis, mechanistic studies and the exploration of new chemical space.

Most reported approaches to NMR mixture analysis rely on comparing and fitting mixture spectra against libraries of spectra of individual compounds. However, experimental uncertainties such as small chemical-shift variations, peak drift and inconsistently measured J-coupling constants complicate direct spectral comparison. Library-based approaches are also restricted to compounds with available reference spectra, which limits their use in the discovery of new chemical entities. Database-independent strategies for mixture analysis are therefore highly desirable.

When a mixture is well controlled and multiple spectra of it can be recorded at varying component concentrations, statistical blind source separation techniques can be applied. Independent component analysis (ICA)~\citep{r5,r6,r7,r8}, sparse component analysis (SCA)~\citep{r9} and non-negative matrix factorisation (NMF)~\citep{r10,r11,r12} have decomposed mixture spectra successfully under such conditions. These approaches exploit covariance patterns across repeated measurements and therefore require several related spectra rather than a single spectrum of an unknown mixture. For example, ICA has resolved mixtures of polyols (glycerol, 1,3-butanediol, 1,2-propanediol, 1,3-propanediol, ethylene glycol and diethylene glycol), enabling quantitative analysis with concentration errors below 10\%, and has been applied to highly overlapping sugar signals in $^1$H-NMR spectra of honey and of food products such as energy drinks and milk~\citep{r8}.

In practice, however, obtaining multiple spectra of the same mixture at controlled and informative concentration ratios is rarely feasible, particularly when little is known about the system or when exploring new chemical space. Component identities, relative abundances and even mixture stability may be unknown, which makes the deliberate generation of suitable measurement series difficult or impossible. As an alternative, several studies have combined orthogonal spectroscopic modalities to improve mixture analysis~\citep{r13,r14,r15}. Complementary techniques provide structural information that is not available from NMR alone and, when chromatographic methods are included, can physically separate the components. Likewise, the joint use of multiple NMR experiments has been reported to improve mixture analysis; for example, combined $^1$H and $^{13}$C data (including DEPT experiments) have been used to infer the functional groups present in mixtures~\citep{r16,r17}.

Although $^1$H-NMR spectra carry a wealth of information in multiplet patterns and J-coupling constants, $^{13}$C-NMR spectra are often easier to interpret because of their much reduced peak overlap, and are therefore frequently preferred in computational approaches. $^1$H-NMR spectra become even harder to analyse in mixtures, especially when multiplets cannot be reliably resolved and little prior knowledge of the composition is available, so that the identification problem becomes ill-defined. When partial information about the components is available, line-fitting methods~\citep{r18,r19}, such as indirect hard modelling, or deep learning can provide partial solutions. For instance, when the structures of the components are known and their spectra can be measured~\citep{r20} or simulated with DFT~\citep{r21}, successful \emph{in situ} reaction analysis has been reported.

Mixture NMR spectra have also been analysed successfully in metabolomics. Because the number of possible metabolites is relatively limited, deep learning models such as NMRVPF~\citep{r22}, NMRformer~\citep{r23} and NMRQNet~\citep{r24} can identify metabolites in mixtures even when concentrations vary by an order of magnitude. A major advantage of a restricted metabolite space is the availability of pure-compound spectra, which reduces the task to compound recognition, similar to database matching, rather than true mixture decomposition. It also enables heuristic and line-fitting approaches~\citep{r25,r26,r27,r28}. For example, a pseudo-Siamese neural network has been reported that detects compounds from a predefined library in $^1$H-NMR spectra of mixtures~\citep{r26}, and similar approaches have been applied to flavour~\citep{r29,r30} and hydrocarbon mixtures~\citep{r31}.

Extending mixture analysis beyond restricted metabolite or compound libraries introduces an important intermediate challenge: separating the resonances of a mixture into groups that are likely to originate from the same molecular component. In such settings the identities of the components are typically unknown, which makes mixture decomposition itself a major challenge. In $^1$H-NMR mixture spectra, individual peaks can be viewed as the elements of an unordered set, and the molecular components as the subsets to which they belong. The permutation-invariant nature of this problem makes it well suited to soft clustering with slot attention, a mechanism originally developed for object-centric representation learning~\citep{r32}. Coherent groupings of input elements (peaks), referred to as slots, emerge without explicit supervision, which allows the input to be decomposed into latent components according to their consistency with the observed data. The competitive nature of slot attention allows peaks to be softly assigned to candidate components and iteratively reassigned according to spectral context. The resulting slots are predicted groupings of peaks rather than identified molecular structures, and thus provide a demixing step that requires neither predefined molecular libraries nor prior assumptions about mixture composition.

Robust and general methods for analysing $^1$H-NMR spectra of chemical mixtures therefore remain highly desirable, despite advances in mixture separation with other spectroscopic modalities~\citep{r33,r34}. Mixture interpretation can be divided into two sequential tasks: demixing the mixture spectrum into component spectra, and interpreting each component spectrum to infer the corresponding molecular structure. Reliable solutions to the first task would benefit research across chemistry, materials science and the life sciences. Here we present the NMR Informed Mixture Identification System (\nimis), a structure-agnostic, chemistry-informed deep learning framework for demixing 1D $^1$H-NMR spectra of mixtures. Unlike approaches that rely on spectral matching or predefined component spaces, \nimis{} formulates mixture separation as a peak-level grouping problem and combines chemically informed attention with slot-based clustering to group mixture peaks by their component of origin. It operates directly on spectral peak information, without spectral libraries or predefined molecular structures, and targets mixtures of moderate spectral complexity.

\section{Methods}

\subsection{NMR mixture dataset}

Experimental NMR spectra of chemical mixtures are scarce. Good practice generally discourages reporting spectra of unresolved mixtures, so fully assigned mixture spectra are rarely available in the literature. Generating such datasets experimentally would require substantial synthetic and analytical effort while covering only a small fraction of the chemical space covered by the large number of spectra available for individual compounds.

Under idealised conditions, without intermolecular interactions and measurement noise, the free-induction decay recorded in a single NMR acquisition is the linear combination of the signals of the individual mixture components, and the Fourier transform preserves this linearity in the frequency domain. For mixtures of small organic molecules at moderate concentrations measured on high-resolution instruments, these conditions are approximately met, so that spectral additivity is a reasonable assumption and mixture spectra can be approximated as linear combinations of the spectra of their components. Given a sufficiently large collection of experimental spectra of isolated compounds, synthetic mixture spectra can therefore be generated \emph{in silico}. This approach preserves the experimental variability of the individual spectra, including measurement noise and reporting inconsistencies, and thereby introduces realistic imperfections into the generated mixtures.

We constructed the dataset from the NMRexp database~\citep{r35}, which contains 3.3 million experimental NMR spectra extracted from the Supporting Information of peer-reviewed publications from 2010 to 2024. To restrict the chemical space to realistic small-molecule chemistry, we considered only $^1$H-NMR spectra of molecules with up to 20 heavy atoms composed of C, H, N, O, S, F, Cl, Br and I. Records lacking a valid SMILES string, solvent, acquisition frequency or processed spectrum were excluded, as were spectra that could not be parsed or that contained inconsistencies between multiplicity and reported J-coupling annotations.

To simulate realistic mixture conditions, all components of a mixture were drawn from the same solvent and spectrometer frequency. Mixtures were generated by randomly selecting 2, 3, 4 or 5 compounds from the available spectra. Each reported resonance of a component is represented by the midpoint of its chemical-shift interval, rounded to 0.01 ppm, and its integral is multiplied by a component proportion sampled from a Dirichlet distribution. The proportions sum to one in each mixture, and no component contributes less than 1\%, which mimics experimental conditions in which very weak signals are absorbed by the baseline noise. Up to three J-coupling constants are retained for each peak, rounded to 0.1 Hz, sorted in descending order, zero-padded and accompanied by a binary mask.

For each mixture, the peak lists, integrals and J-coupling values of the selected components are concatenated and sorted by chemical shift, so that the peaks form an unordered set without positional bias. Because real peak profiles are not available from the processed data, closely overlapping resonances from different components are retained as separate peaks rather than merged into composite signals during mixture generation. This is reasonable for mixtures of moderate spectral complexity, in which substantial portions of the spectra are resolved, but may limit applicability to more congested experimental spectra.

Data are partitioned at the molecular level before mixture construction to prevent leakage of molecular identity among the training, validation and test sets. The training set contains 158,005 mixtures comprising 65,809 unique molecules and 9,742 unique Bemis--Murcko scaffolds. The molecule-holdout validation and test sets each contain 16,000 mixtures: the validation set comprises 1,370 unique molecules and 638 scaffolds, and the test set 1,372 unique molecules and 633 scaffolds. No molecules are shared between the training set and the held-out sets, although 501 scaffolds are shared between the training and test sets, corresponding to 79.15\% of the scaffolds represented in the test set. Although the model does not receive molecular structures, this partitioning evaluates its ability to generalise to unseen compounds while retaining partial coverage of the chemical frameworks encountered in training. During construction of the training set, molecules belonging to scaffolds represented fewer than 15 times are preferentially sampled, and mixtures containing scaffolds that still fail to reach this threshold are removed. This procedure provides broad scaffold coverage while ensuring adequate representation of every retained scaffold.

To ensure representation of low-abundance components, 11.39\%, 12.75\%, 11.26\%, 9.85\% and 54.75\% of component occurrences in the final training set fall within the relative concentration ranges 1--5\%, 5--10\%, 10--15\%, 15--20\% and above 20\%, respectively; the validation and test sets have comparable distributions. To represent experimental uncertainty, multiplicative Gaussian perturbations with a standard deviation of 10\% are applied independently to the peak integrals during training. This augmentation approximates integration variability arising from spectral processing, reduced signal-to-noise ratios and moderate deviations from quantitative acquisition conditions, such as intensity distortions caused by incomplete longitudinal relaxation when repetition times are shorter than those required for quantitative NMR, particularly in rapid reaction monitoring. Because the source spectra come from a large literature-derived database spanning many years, a realistic correction for incomplete relaxation is impossible. Random integral perturbation is therefore a reasonable approximation of the associated uncertainty, and it prevents the model from treating the reported integrals as exact quantitative measurements.

\subsection{Model architecture}

\begin{figure}[!htbp]
\centering
\figimg{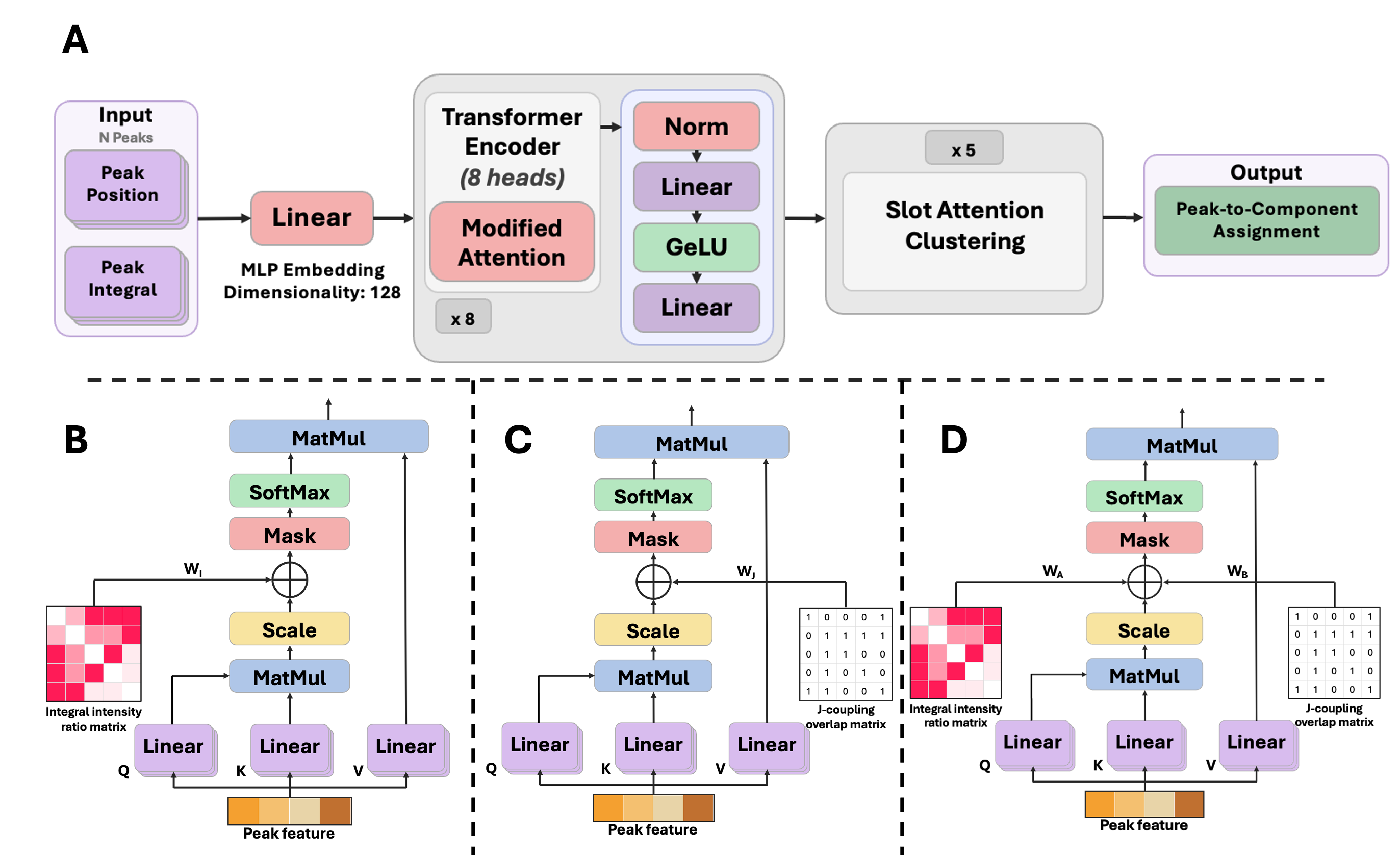}
\caption{\textbf{A.}~Overall \nimis{} architecture. \textbf{B.}~Attention modified by an integral-ratio matrix. \textbf{C.}~Attention modified by a J-coupling matching matrix. \textbf{D.}~Attention modified by both an integral-ratio matrix and a J-coupling matching matrix.}
\label{fig:arch}
\end{figure}

\nimis{} is a two-stage pipeline (Figure~\ref{fig:arch}A) in which individual spectral peaks are first encoded into contextualised representations and then grouped into mixture components by slot-based clustering. The input is a processed $^1$H-NMR peak list containing peak positions, integral values and J-coupling constants. Each peak can carry up to three J-coupling constants; more complex peaks are treated as multiplets. This peak-level information can be obtained manually or with standard NMR processing software. This upstream step is outside the scope of \nimis{} and of the present work.

Two scalar features are derived for each peak: its chemical shift (in ppm) and the integral value obtained in the processing step. The chemical shift is encoded with Fourier features, i.e.\ sine and cosine functions at 16 geometrically spaced periods between 16 and 0.01 ppm, which allow the model to resolve both coarse spectral position and closely spaced peaks. The integral is log-transformed to increase sensitivity to small integrals, such as those of low-abundance components. The resulting 33 features are embedded into a 128-dimensional representation by a three-layer MLP, and a Transformer encoder then contextualises each resonance relative to all other peaks in the mixture. We consider three chemically informed adaptations of scaled dot-product attention (Figure~\ref{fig:arch}B--D). They exploit two complementary sources of information in $^1$H-NMR spectra: relative peak integrals, which carry information about proton count and component abundance, and J-coupling relationships, which reflect through-bond spin--spin interactions between nuclei. The corresponding pairwise relationships are represented by an integral-ratio matrix $C_I$ and a J-coupling matching matrix $C_J$.

The J-coupling matching matrix is a binary pairwise matrix. An entry $C_{J,ij}$ is set to 1 when peaks $i$ and $j$ share at least one reported J-coupling value within a tolerance of 0.1~Hz, and to 0 otherwise. The matrix thus provides a sparse indication that two resonances may originate from the same coupled spin system. Incorporating it into the attention mechanism allows coupling information to influence the contextualisation of peaks without requiring assignments to be determined solely from J-value matches. The integral-ratio matrix is constructed from the pairwise ratios of the reported peak integrals. For peaks $i$ and $j$ with integrals $I_i$ and $I_j$, the corresponding entry is
\begin{equation}
C_{I,ij} = \min\!\left(\frac{I_i}{I_j},\,\frac{I_j}{I_i}\right).
\end{equation}
Ratios greater than 1 are thereby inverted, which restricts all entries to the interval $[0,1]$. Values near 1 indicate peaks with similar integrals, and values near 0 indicate a large difference in relative intensity. Ratios involving a zero-valued integral are set to 0, and diagonal entries are set to 1.

The encoder is a stack of Transformer layers whose attention directly incorporates peak integral and J-coupling information:
\begin{equation}
\mathrm{Attention}(Q,K,V) = \mathrm{softmax}\!\left(\frac{QK^{\top}}{\sqrt{d_k}} + W_I C_I + W_J C_J\right)V,
\end{equation}
where $Q$ and $K$ are the query and key matrices, $d_k$ is the dimensionality of the key vectors, and $W_I$ and $W_J$ are learnable scalar weights that control the influence of the integral and J-coupling pairwise matrices, respectively. Introducing these relationships at the level of the attention logits allows the model to strengthen or suppress interactions between peaks according to their pairwise spectral consistency, while retaining the flexibility to adjust the contribution of each source of information during training. Three otherwise identical attention formulations are evaluated: attention incorporating only the J-coupling matching matrix (Figure~\ref{fig:arch}C), only the integral-ratio matrix (Figure~\ref{fig:arch}B), or both matrices (Figure~\ref{fig:arch}D).

The contextualised peak embeddings are then grouped into latent mixture components by slot attention, which assigns peaks to slots iteratively. The output of \nimis{} is a peak-level assignment table containing peak positions, arbitrary predicted component labels and assignment confidence values, referred to as \nimis{} scores. The \nimis{} score is the post-softmax probability that the model assigns to the predicted component after optimal slot alignment with the Hungarian algorithm. Scores close to 1 indicate confident peak-to-component assignments, whereas lower scores reflect greater ambiguity. Peaks assigned to the same component are grouped together, and the \nimis{} score provides a direct measure of prediction confidence. Details of the model architecture and training are provided in the Supporting Information Section 1.

\subsection{Training objective}

The model is trained to assign peaks to their mixture components with a cross-entropy loss on the peak-to-slot attention probabilities. Because the ordering of the predicted slots is arbitrary and need not match the ordering of the ground-truth components, Hungarian matching is used to resolve the permutation ambiguity. For each mixture, a cost matrix between predicted slots and ground-truth components is built from the negative log-probabilities that the slots assign to the peaks of each component, and the Hungarian algorithm finds the slot-to-component assignment of minimum total cost. The predicted slot distributions are then realigned to this assignment, and the final loss is computed after realignment, which makes training permutation-invariant.

\section{Results and discussion}

We trained the three chemically informed configurations and a conventional Transformer, and evaluated all four at four levels of integral perturbation. As shown in Table~\ref{tab:noise}, all chemically informed configurations are more accurate than conventional attention at every noise level. The J-coupling-only configuration is the most accurate throughout, decreasing from 82.28\% without noise to 72.44\% at 15\% added integral noise, compared with 79.54\% and 69.99\%, respectively, for conventional attention. The combined-prior and integral-ratio-only configurations also outperform conventional attention at every noise level, which suggests that integral ratios carry meaningful information about peak-to-component attribution. However, the consistently higher accuracy of the J-coupling-only configuration relative to the combined-prior configuration suggests that the two priors may sometimes provide conflicting grouping cues, which limits the benefit of using both.

\begin{table}[!htbp]
\centering
\small
\caption{Peak-to-component assignment accuracy (\%) of the three chemically informed attention configurations and the vanilla Transformer at four levels of added integral noise, evaluated on the molecule-disjoint test set. The best value in each row is in bold.}
\label{tab:noise}
\begin{tabular}{ccccc}
\toprule
Added integral & Both chemical & J-coupling matching & Integral-ratio matrix & Vanilla\\
noise (\%) & priors & matrix only & only & Transformer\\
\midrule
0  & 81.89 & \textbf{82.28} & 81.92 & 79.54\\
5  & 80.55 & \textbf{81.16} & 80.65 & 78.58\\
10 & 76.57 & \textbf{77.34} & 76.44 & 74.61\\
15 & 71.71 & \textbf{72.44} & 71.31 & 69.99\\
\bottomrule
\end{tabular}
\end{table}

Table~\ref{tab:fraction} breaks the accuracy down by component fraction. The J-coupling-only configuration is the most accurate in every abundance range except 1--5\%, where the integral-ratio-only configuration performs best. This exception may reflect the comparatively small integrals of low-abundance components, which can be a distinctive cue for grouping their peaks, and may be most informative in simpler mixtures in which a minor component is accompanied by components at substantially higher concentrations. Evaluating components at 1--5\% abundance, however, may be a less realistic application scenario, because their weak signals are difficult to distinguish from the baseline, which hampers reliable peak identification and integration.

\begin{table}[!htbp]
\centering
\small
\caption{Peak-to-component assignment accuracy (\%) by component fraction at 10\% added integral noise. The best value in each row is in bold.}
\label{tab:fraction}
\begin{tabular}{cccccc}
\toprule
Component & Number & Both chemical & J-coupling matching & Integral-ratio matrix & Vanilla\\
fraction (\%) & of peaks & priors & matrix only & only & Transformer\\
\midrule
1--5   & 44,443  & 86.70 & 86.55 & \textbf{87.18} & 84.97\\
5--10  & 49,739  & 75.66 & \textbf{76.52} & 76.07 & 73.60\\
10--15 & 43,635  & 69.35 & \textbf{69.86} & 68.65 & 66.77\\
15--20 & 37,985  & 66.82 & \textbf{67.41} & 65.84 & 63.36\\
$>$20  & 212,844 & 77.88 & \textbf{78.91} & 77.78 & 76.30\\
\bottomrule
\end{tabular}
\end{table}

Integral relationships can be informative within a known compound, but pairwise ratios in an unknown mixture depend on proton count, component abundance, integration uncertainty and acquisition conditions. The lower accuracy of the integral-ratio configuration relative to the J-coupling-only configuration may therefore arise because similar ratios occur between unrelated peaks, particularly between low-abundance components and unrelated resonances representing fewer proton environments. Such non-specific relationships introduce an ambiguous attention bias and reduce assignment accuracy. In contrast, the J-coupling matrix is sparser but more chemically specific, although it connects only peaks with matching reported coupling constants. Because the J-coupling-only configuration was the most accurate at every noise level and, except for the 1--5\% range, at every concentration range, the final \nimis{} architecture incorporates only the J-coupling attention bias (Figure~\ref{fig:arch}C).

\begin{figure}[!htbp]
\centering
\figimg{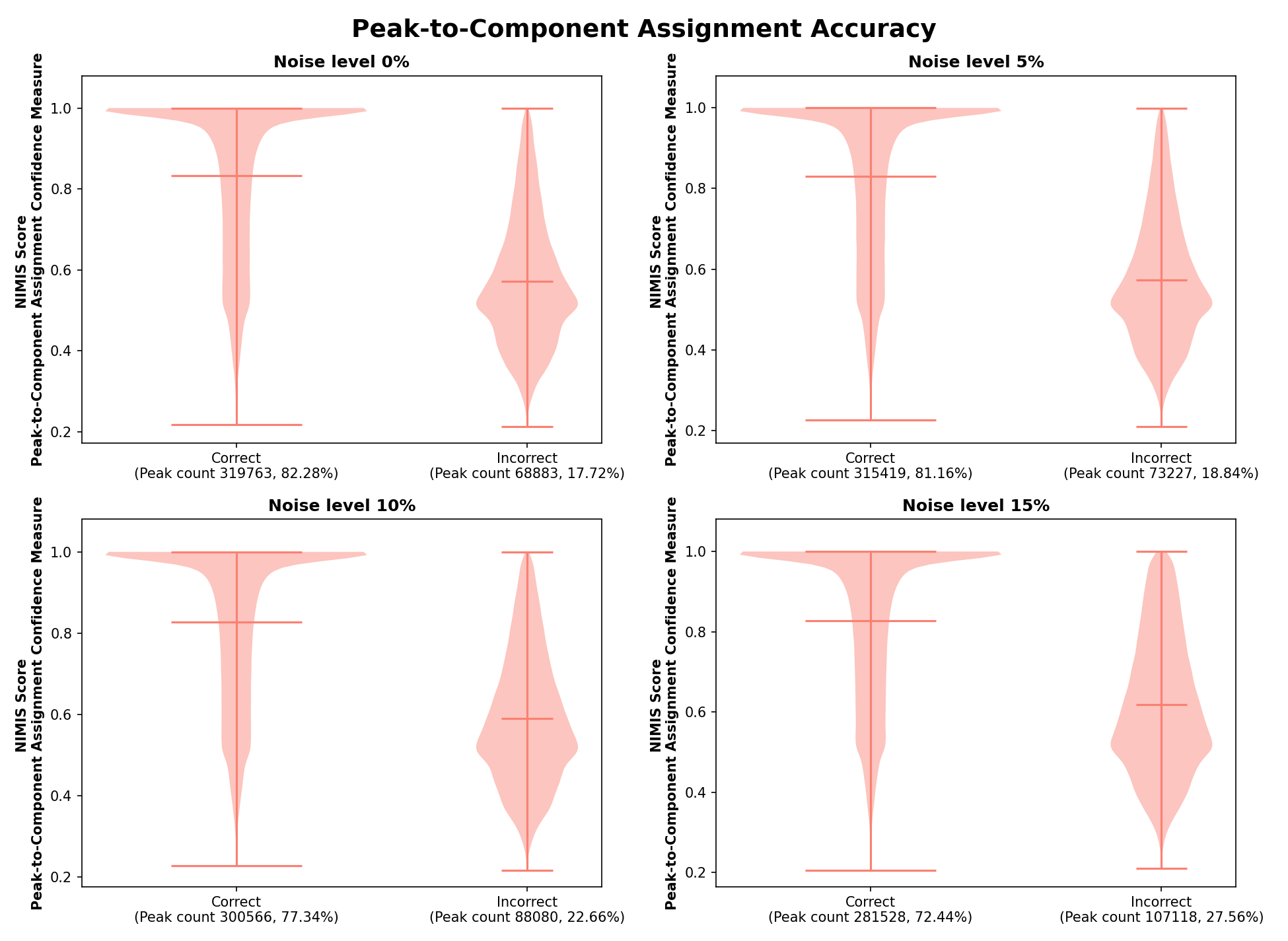}
\caption{Confidence of correct and incorrect peak-to-component assignments at four levels of added integral noise: 0\% (A), 5\% (B), 10\% (C) and 15\% (D). Violins show the distribution of \nimis{} scores; horizontal bars mark the mean and extrema.}
\label{fig:violin}
\end{figure}

At 10\% random noise in the peak integrals, the model correctly assigns 77.34\% of peaks after 300 training epochs (Figure~\ref{fig:violin}C). Correctly assigned peaks have a high mean \nimis{} score of 0.828 (SD $=0.186$), and 60.1\% of them score above this mean. Incorrectly assigned peaks have a markedly lower mean score of 0.590 (SD $=0.158$). The score therefore discriminates between correct and incorrect assignments and can help to identify the most reliable peak-to-component assignments in practical mixture analysis.

Although the model was trained with 10\% random perturbation of the peak integrals to represent experimental uncertainty, we also tested it at perturbation levels from 0 to 15\% to assess its applicability to noisier experiments. The mean score of correctly assigned peaks remains essentially unchanged across these levels (Figure~\ref{fig:violin}A, B and D), whereas the mean score of incorrectly assigned peaks increases from 0.571 without added perturbation to 0.618 at 15\%, and the fraction of incorrectly assigned peaks increases from 17.72\% to 27.56\%. Increasing perturbation thus introduces additional errors and modestly narrows the confidence separation between correct and incorrect predictions, but accuracy declines only gradually, from 82.28\% to 72.44\%. These results demonstrate tolerance to moderate integration uncertainty. They also indicate that accurate peak integrals, obtained with sufficient signal averaging, appropriate relaxation delays and careful integration, would further improve peak grouping and assignment reliability.

Random perturbation of the integrals models uncorrelated integration noise. To assess sensitivity to systematic errors, we additionally applied a uniform multiplicative bias of $\pm5$, $\pm10$ and $\pm15\%$ to every integral, every J-coupling constant, or both, on top of the 10\% random integral noise used in training (Table~\ref{tab:bias}). The J-coupling matching matrix was recomputed from the perturbed J-values with the standard 0.1~Hz tolerance. Accuracy decreased smoothly and without abrupt failure as the bias in the integrals grew: from 77.30\% for unperturbed inputs to 76.87--77.05\% at $\pm5\%$, 75.81--76.10\% at $\pm10\%$ and 74.29--74.77\% at $\pm15\%$, and it was slightly lower for under- than for over-estimated integrals. A uniform bias in the J-coupling constants had essentially no effect (accuracy between 77.16\% and 77.42\% at all levels), as expected, because scaling all values by the same factor shifts both members of a matching pair equally and thus preserves most matches. Biasing integrals and J-couplings together gave accuracies close to those obtained by biasing the integrals alone (74.36--74.60\% at $\pm15\%$), indicating no additional sensitivity to the J-couplings. The vanilla Transformer, which does not use J-couplings, was similarly sensitive to integral bias (72.30--72.48\% at $\pm15\%$) and was less accurate than \nimis{} at every bias level. The loss of accuracy therefore originates from the integral input itself rather than from the chemically informed attention, and the advantage conferred by the J-coupling prior is retained under systematic integral error. For integral biases of up to $\pm10\%$, \nimis{} accuracy remains above 75.8\%, although accurately calibrated integrals remain desirable.

\begin{table}[!htbp]
\centering
\small
\caption{Peak-to-component assignment accuracy (\%) on the molecule-disjoint test set when a uniform bias is applied to the input integrals and/or J-coupling constants, in addition to the random 10\% integral noise used in training. Negative bias values scale all affected inputs down and positive values scale them up. Values are means over five evaluation seeds (standard deviation at most 0.08). The vanilla Transformer does not use J-couplings, so only integral bias applies.}
\label{tab:bias}
\begin{tabular}{lcccc}
\toprule
 & \multicolumn{3}{c}{\nimis} & Vanilla Transformer\\
\cmidrule(lr){2-4}\cmidrule(lr){5-5}
Uniform bias & Integrals & J-couplings & Integrals + J-couplings & Integrals\\
\midrule
None        & 77.30 & 77.30 & 77.30 & 74.69\\
$-15\%$     & 74.29 & 77.42 & 74.36 & 72.30\\
$-10\%$     & 75.81 & 77.41 & 75.90 & 73.58\\
$-5\%$      & 76.87 & 77.40 & 76.97 & 74.39\\
$+5\%$      & 77.05 & 77.18 & 76.87 & 74.46\\
$+10\%$     & 76.10 & 77.16 & 75.95 & 73.68\\
$+15\%$     & 74.77 & 77.20 & 74.60 & 72.48\\
\bottomrule
\end{tabular}
\end{table}

Given the benefit of the J-coupling matching matrix, we evaluated the effect of the fixed 0.1~Hz matching tolerance. J-coupling constants in Hz are absolute, but they are measured in practice as peak separations in ppm, which depend on the operating frequency of the spectrometer. The precision of reported J-values is also influenced by spectral resolution, linewidth, signal-to-noise ratio, peak overlap and reporting precision. The tolerance therefore determines the balance between recovering genuine coupling relationships and introducing spurious matches, and as it increases, more unrelated peaks may be connected (Table~\ref{tab:tol}). Increasing the tolerance from 0.00 to 1.00~Hz raised the fraction of peaks with a match from 22.04\% to 39.73\%, whereas accuracy changed only from 77.24\% to 77.02\%. Model performance was therefore largely insensitive to a substantial increase in the matching rate, which suggests that the attention mechanism can prioritise genuine coupling relationships over spurious matches. The learned attention weights assigned to true and false matches at different tolerances (Table~\ref{tab:attn}) support this interpretation.

\begin{table}[!htbp]
\centering
\small
\caption{Effect of the J-coupling matching tolerance on assignment accuracy.}
\label{tab:tol}
\begin{tabular}{ccc}
\toprule
Tolerance (Hz) & Accuracy (\%) & Peaks matched (\%)\\
\midrule
0.00 & 77.24 & 22.04\\
0.10 & 77.40 & 26.68\\
0.20 & 77.42 & 31.70\\
0.30 & 77.42 & 33.64\\
0.50 & 77.31 & 37.37\\
1.00 & 77.02 & 39.73\\
\bottomrule
\end{tabular}
\end{table}

At every tolerance, genuine matches received approximately twice as much attention as false matches, which shows that the model selectively attends to peaks connected by meaningful J-coupling relationships. Among peak pairs not connected by the coupling criterion, peaks from the same compound still received more attention than peaks from different compounds. Their attention weights were, however, comparable to those of false matches at larger tolerances. \nimis{} therefore combines the sparse J-coupling prior with the broader learned spectral context, which allows genuine coupling relationships to be prioritised without requiring all peaks of a component to be connected through reported J-values.

\begin{table}[!htbp]
\centering
\small
\caption{Mean attention weight between peak pairs by match type and tolerance.}
\label{tab:attn}
\begin{tabular}{lcc}
\toprule
Match type & Tolerance (Hz) & Mean attention\\
\midrule
True match & 0.1 & 0.0891\\
False match & 0.1 & 0.0387\\
True match & 1.0 & 0.0749\\
False match & 1.0 & 0.0395\\
Unmatched (same component) & -- & 0.0450\\
Unmatched (different components) & -- & 0.0259\\
\bottomrule
\end{tabular}
\end{table}

To test whether J-couplings alone can reproduce this performance, peaks matched at the 0.1~Hz tolerance were grouped by constructing an adjacency matrix, and the resulting connected components were aligned with the ground-truth components with the Hungarian algorithm. This deterministic approach recovers only 24.20\% of peak assignments, below the random-assignment baseline of 43.79\% (Table~\ref{tab:baselines}). The reason is that only 43.5\% of the peaks in the test set carry any reported J-coupling constant and only 26.68\% have a match at 0.1~Hz (Table~\ref{tab:tol}). The remaining peaks, which include all singlets, form isolated clusters, so the number of predicted groups far exceeds the number of mixture components. J-coupling matching alone is therefore not conclusive for most peaks. To give the coupling information a fair test, we extended this heuristic into three further baselines, each constrained to the known number of components. In J-coupling with random completion, the largest groups of J-connected peaks seed the components and every remaining peak is assigned at random. In integral matching alone, each component is assumed to have a characteristic integral per proton, and peaks are grouped by how well their integrals fit whole-number multiples of that value. In J-coupling with integral matching, J-connected groups seed the components, their per-proton integral values are estimated, and every remaining peak is assigned to the component whose integral pattern fits it best. In addition, peaks were allocated at random among the known number of components. Random assignment groups 43.79\% of peaks correctly, J-coupling with random completion 46.28\%, integral matching alone 58.55\% and the combined J-coupling and integral heuristic 60.88\%, compared with 77.34\% for \nimis{} (Table~\ref{tab:baselines}). All heuristic baselines therefore remain well below \nimis. A Transformer with unbiased attention reaches 74.61\%, which indicates that most of the model's performance arises from learned spectral context, while the chemical prior provides a smaller but consistent additional improvement.

The strong performance of the unbiased Transformer is important because J-coupling information is not consistently available in experimental spectra. Poorly resolved, overlapping or incompletely reported multiplets may contain few or no extractable coupling constants. As fewer J-values are available, the J-coupling matrix becomes sparser and influences a progressively smaller proportion of the attention relationships. If no coupling values are reported, the matrix contains only its diagonal, and the attention reduces to that of the unbiased Transformer up to a constant self-attention term (Supporting Information, Table S6). \nimis{} is therefore expected to degrade gradually towards the conventional Transformer baseline, rather than fail, when coupling information is incomplete. This behaviour is valuable for the intended application, in which processed peak lists may contain varying amounts of coupling information.

To our knowledge, no previously reported model addresses the same structure-agnostic peak-to-component grouping task as \nimis, which precludes a direct comparison with an equivalent method. The closest related approach is NMRformer. Although it identifies metabolites within a predefined space of 71 candidate compounds and receives the complete mixture spectrum alongside the peak positions, it also treats NMR mixture demixing as a peak-to-component assignment problem. NMRformer reports higher assignment accuracies than \nimis, depending on the metabolite class. This difference may partly reflect its richer input: it receives the complete spectral line shape and can extract intensity and multiplet information that is not available from the processed peak lists used by \nimis. NMRformer also operates in a restricted metabolite space of 71 compounds with substantially lower molecular and scaffold diversity. The two models therefore serve different purposes. NMRformer identifies metabolites within a predefined candidate space, whereas \nimis{} performs structure-agnostic peak grouping across a broad and chemically diverse molecular domain without a component library, and thus generates component-level spectral representations of previously unseen molecules.

\begin{table}[!htbp]
\centering
\small
\caption{Peak-to-component assignment accuracy (\%) of \nimis, the vanilla Transformer and the heuristic baselines on the molecule-disjoint test set at 10\% integral noise, alongside the range reported for NMRformer on a different task and dataset. The best value among the methods evaluated on this test set is in bold.}
\label{tab:baselines}
\begin{tabular}{lc}
\toprule
Approach & Accuracy (\%)\\
\midrule
\nimis & \textbf{77.34}\\
Vanilla Transformer & 74.61\\
Random assignment & 43.79\\
J-coupling matching only & 24.20\\
J-coupling with random completion & 46.28\\
Integral matching alone & 58.55\\
J-coupling with integral matching & 60.88\\
NMRformer (as reported) & 87.2--92.5\\
\bottomrule
\end{tabular}
\end{table}

\begin{figure}[!htbp]
\centering
\figimg[0.65\linewidth]{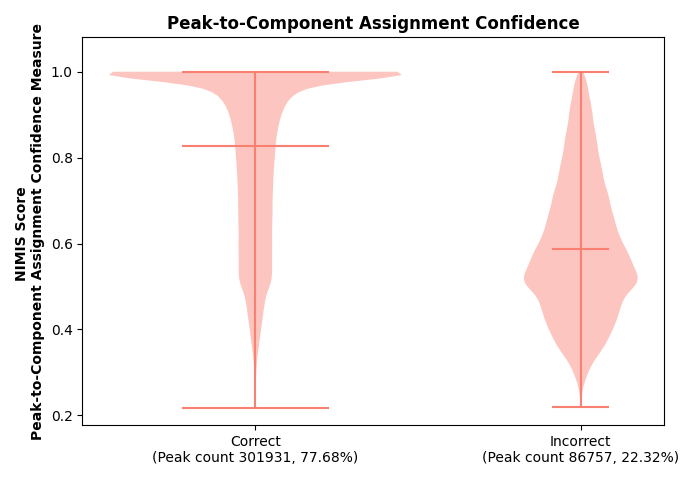}
\caption{Confidence of correct and incorrect assignments on the scaffold-disjoint test set (10\% added integral noise).}
\label{fig:scaffold}
\end{figure}

To further assess generalisation to unseen molecules and molecular frameworks, we constructed a scaffold-disjoint test set of 15,991 mixtures containing 13,113 unique molecules and 1,960 Bemis--Murcko scaffolds. Despite this more stringent separation, \nimis{} correctly assigns 77.68\% of peaks, compared with 77.34\% on the molecule-holdout set, and preserves a clear separation between the confidence distributions of correct and incorrect assignments (Figure~\ref{fig:scaffold}). This essentially unchanged performance indicates that \nimis{} does not depend on memorising molecular identities or common scaffolds, and that its peak-grouping strategy generalises to unseen regions of chemical space. Although its accuracy is lower than that reported for NMRformer, the two models offer complementary capabilities: NMRformer achieves high accuracy within a defined metabolite space using rich spectral inputs, whereas \nimis{} enables structure-agnostic, component-level peak grouping across unseen molecules and scaffolds without a predefined component library.

\subsection{Assignment interpretation}

\begin{figure}[!htbp]
\centering
\figimg{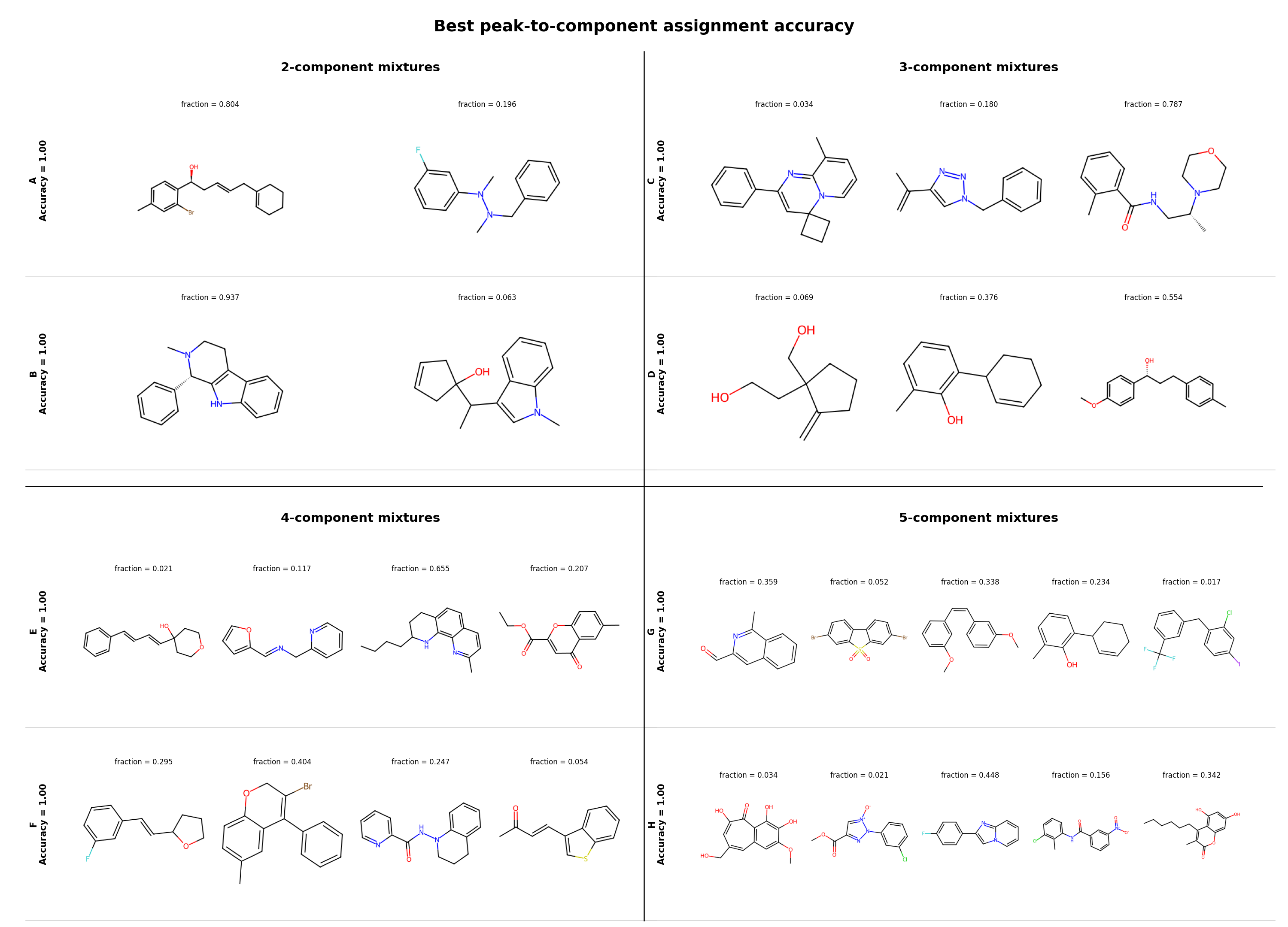}
\caption{Best-performing simulated mixtures: the two most accurately assigned two-, three-, four- and five-component mixtures. Component fractions are shown above each structure. Molecular structures are not available to \nimis{} during prediction.}
\label{fig:best}
\end{figure}

\begin{figure}[!htbp]
\centering
\figimg{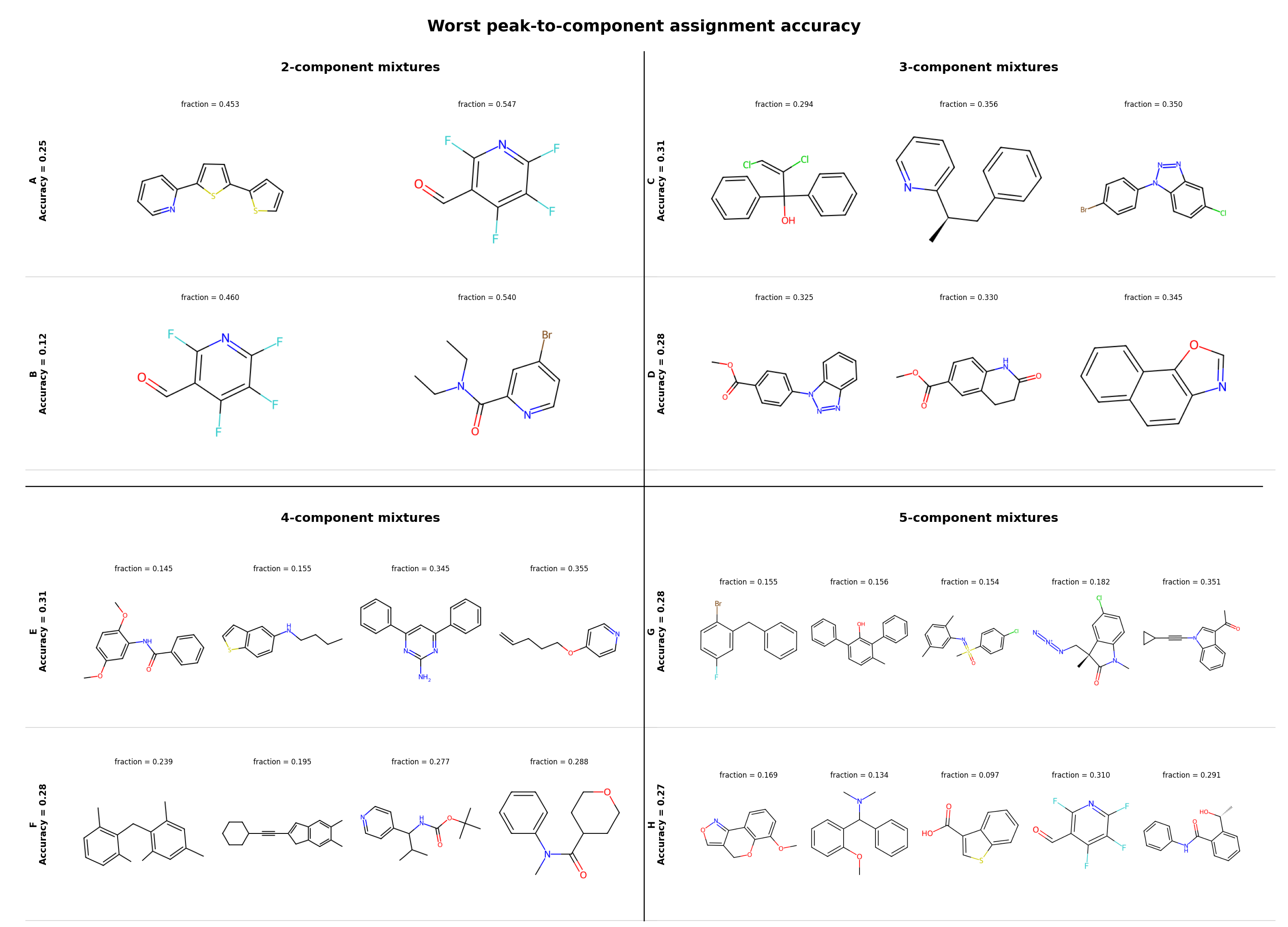}
\caption{Worst-performing simulated mixtures: the two least accurately assigned two-, three-, four- and five-component mixtures. Component fractions are shown above each structure. Molecular structures are not available to \nimis{} during prediction.}
\label{fig:worst}
\end{figure}

To examine the conditions associated with successful and unsuccessful demixing, we inspected the two best and two worst performing two-, three-, four- and five-component mixtures (Figures~\ref{fig:best} and~\ref{fig:worst}). Molecular structures are not available to \nimis{} during prediction. They are shown to determine whether assignment performance is associated with specific structural features or levels of molecular complexity. At every complexity level, some mixtures are assigned perfectly, with every peak attributed to its component of origin (Figure~\ref{fig:best}A--H). These mixtures are chemically diverse, containing mostly aromatic and some aliphatic systems, heterocycles, halogenated compounds, and a range of oxygen-, nitrogen-, fluorine- and sulphur-containing functional groups. High accuracy is also retained across a broad range of component abundances, including fractions close to the minimum represented in training.

The worst-performing mixtures (Figure~\ref{fig:worst}) share no single structural feature or level of molecular complexity that consistently distinguishes unsuccessful assignments. They differ from the best-performing mixtures in component abundance: every best-performing example contains at least one clearly differentiated component fraction, whereas all worst-performing examples contain components at comparable abundances. This suggests that assignment becomes harder when similar spectral environments occur in several components at comparable abundances, particularly when the corresponding peaks also lack discriminative J-coupling relationships.

One recurring worst-case molecule is 2,4,5,6-tetrafluoronicotinaldehyde, which appears in three of the eight lowest-accuracy mixtures (Figure~\ref{fig:worst}A, B and H). Despite its 12 heavy atoms, it has a single proton and therefore contributes only one peak to the mixture spectrum. Assigning a lone peak to its component without any intra-component spectral information is inherently ill-posed, so poor performance on such mixtures reflects a limitation of the input rather than of the model. Molecules of this kind occur throughout the dataset and cannot be controlled for at the sample level, so they inevitably appear among the unsuccessful cases; we do not interpret them as a further weakness of \nimis.

\begin{figure}[!htbp]
\centering
\figimg[0.8\linewidth]{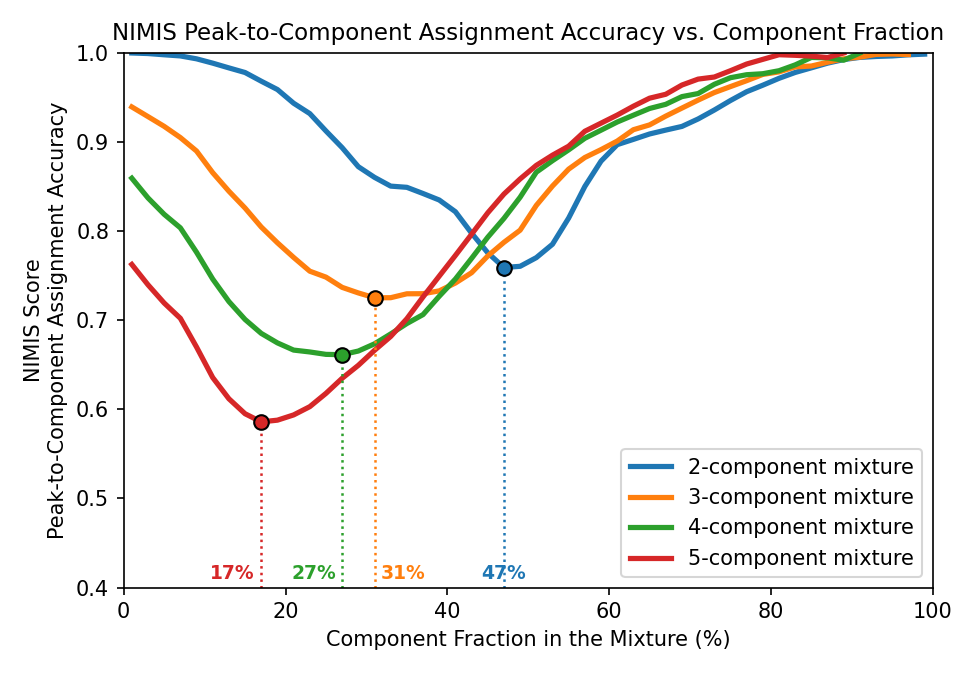}
\caption{Peak-to-component assignment accuracy as a function of component fraction. Accuracy of \nimis{} on the molecule-disjoint test set at 10\% added integral noise, stratified by the number of mixture components.}
\label{fig:fraction}
\end{figure}

To test the abundance hypothesis, we evaluated assignment accuracy as a function of component fraction, stratified by the number of mixture components (Figure~\ref{fig:fraction}). For each mixture size, accuracy is lowest when component abundances approach a uniform distribution: the minima for two-, three-, four- and five-component mixtures occur at component fractions of 47\%, 31\%, 27\% and 17\%, close to the corresponding equimolar fractions. Relative abundance therefore appears to be a useful grouping cue when components occur at distinct concentrations, and a less discriminative one in approximately equimolar mixtures. This also provides context for the lower accuracy observed when the integral-ratio attention bias is incorporated (Table~\ref{tab:fraction}): similar integral ratios occur between unrelated peaks from different components, which increases the number of ambiguous or spurious relationships presented to the model. When component abundances are similar and J-coupling information is absent, the processed peak list may contain insufficient evidence to determine component origin unambiguously.

Peak overlap is a further limitation of the current \nimis{} formulation. Each input peak is assigned to a single component, and the model does not explicitly represent composite signals with contributions from two or more molecules. An unresolved composite peak can therefore be attributed to only one mixture component. In the synthetic dataset, coincident or closely spaced resonances from different components remain separate labelled entries, because the experimental lineshapes required to construct realistic composite signals are unavailable. In experimentally acquired spectra, however, sufficiently close resonances may appear as a single unresolved peak or multiplet. The frequency of such ambiguous signals is expected to increase with the number of components and the degree of spectral congestion, potentially contributing to incorrect assignments.

\subsection{Case studies on experimental mixtures}

\begin{figure}[!htbp]
\centering
\figimg{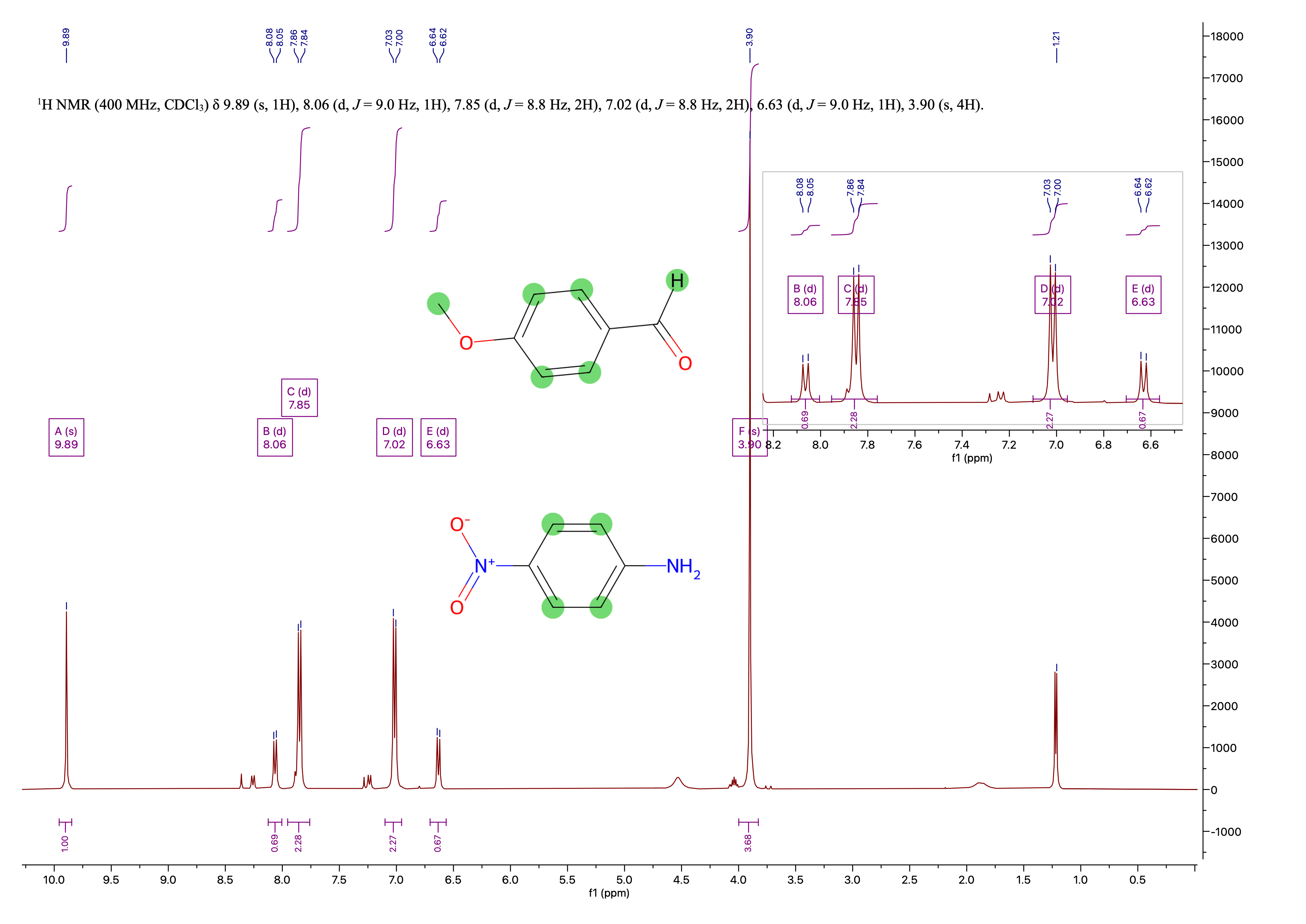}
\caption{Experimental $^1$H-NMR spectrum of a binary mixture of 4-methoxybenzaldehyde and 4-nitroaniline (400~MHz, CDCl$_3$). Green markers indicate the hydrogen-bearing positions of each component that were correctly grouped together. The multiplet report above the spectrum was generated by automatic analysis in MestReNova and lists rounded proton counts; the unrounded integrals shown under each peak were used as \nimis{} input (Supporting Information Table~S7).}
\label{fig:case1}
\end{figure}

\begin{figure}[!htbp]
\centering
\figimg{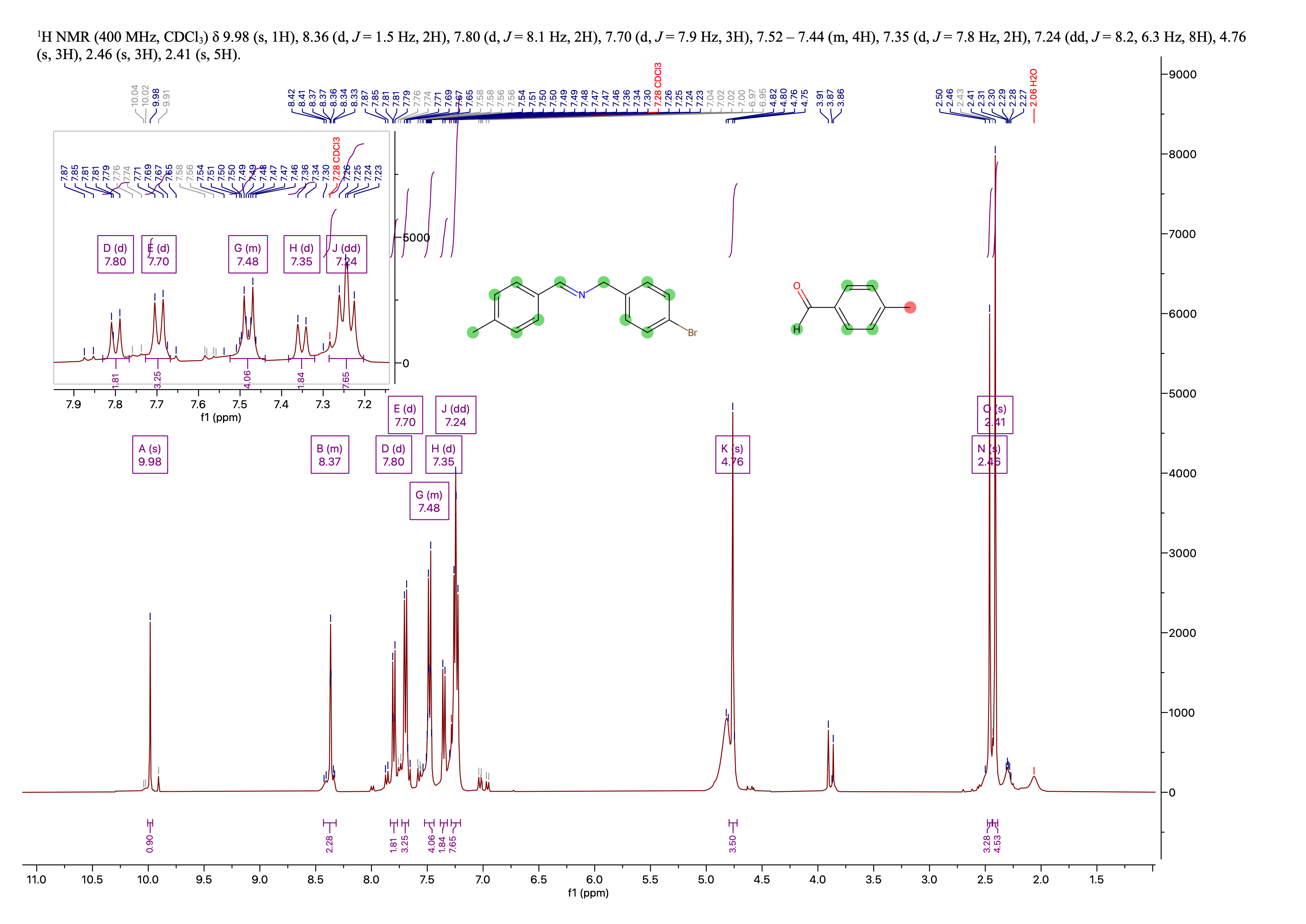}
\caption{Experimental $^1$H-NMR spectrum of a binary mixture of 4-methylbenzaldehyde and (\emph{E})-N-(4-bromobenzyl)-1-(4-methylphenyl)methanimine (400~MHz, CDCl$_3$). Green markers indicate the hydrogen-bearing positions of each component that were correctly grouped together. The multiplet report above the spectrum was generated by automatic analysis in MestReNova and lists rounded proton counts; the unrounded integrals shown under each peak were used as \nimis{} input (Supporting Information Table~S7).}
\label{fig:case2}
\end{figure}

\begin{figure}[!htbp]
\centering
\figimg{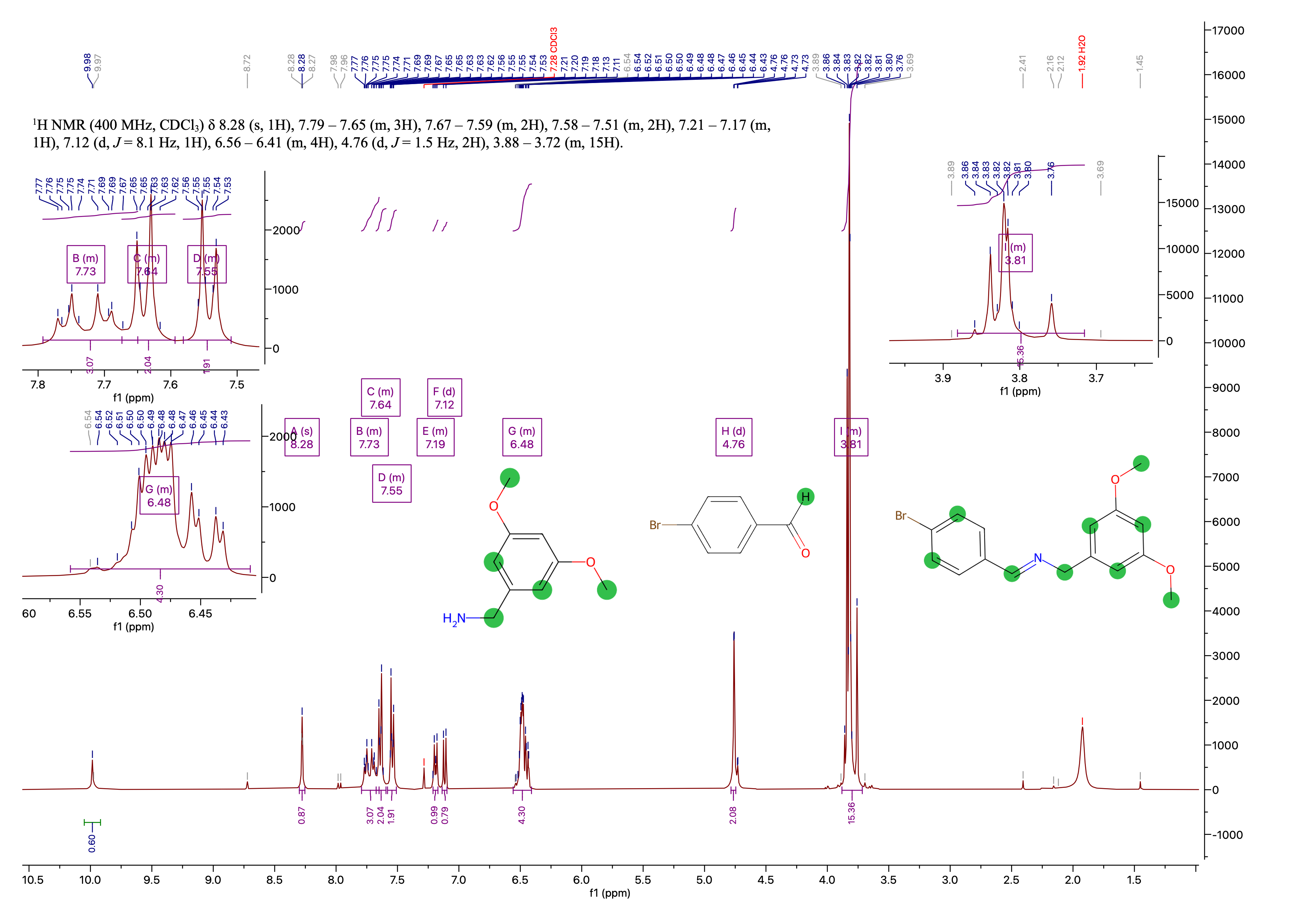}
\caption{Experimental $^1$H-NMR spectrum (400~MHz, CDCl$_3$) of a ternary mixture of 4-bromobenzaldehyde, 3,5-dimethoxybenzylamine and the imine formed from them. The insets show the expanded aromatic and methoxy regions. Green markers indicate the hydrogen-bearing positions of each component that were correctly grouped together. The multiplet report above the spectrum was generated by automatic analysis in MestReNova and lists rounded proton counts; the unrounded integrals shown under each peak were used as \nimis{} input (Supporting Information Table~S7).}
\label{fig:case3}
\end{figure}

Nevertheless, three experimental $^1$H-NMR spectra of mixtures, two binary and one ternary, were available from earlier experimental work of the first author at Imperial College London (see Acknowledgements), which allowed us to assess the practical applicability of \nimis{} beyond the constructed datasets. All spectra were processed with an automated workflow. Signals were annotated with the automatic annotation tool in MestReNova, and the resulting signal information was passed to \nimis{} without manual correction during annotation and component grouping, with the single exception noted for the ternary mixture below. Peak positions, multiplicities and J-coupling constants were taken from the automatic analysis. The multiplet report generated by this analysis, shown above each spectrum in Figures~\ref{fig:case1}--\ref{fig:case3}, gives proton counts rounded to whole numbers. The \nimis{} inputs therefore used the unrounded integral values determined by the same automatic analysis and displayed under each peak, rather than the rounded values of the report. The complete input peak lists are given in Supporting Information Table~S7.

The first example is a binary mixture of 4-methoxybenzaldehyde and 4-nitroaniline (Figure~\ref{fig:case1}). It is a well-suited test case because the components are present in unequal amounts (approximately 3:1 from the integrals), so that relative abundance provides a meaningful cue for grouping peaks, and each component gives clearly resolved signals with well-defined coupling patterns. 4-Methoxybenzaldehyde shows an aldehyde singlet at 9.89~ppm, a methoxy singlet at 3.90~ppm and a pair of aromatic doublets at 7.85 and 7.02~ppm ($J=8.8$~Hz), whereas 4-nitroaniline shows a pair of aromatic doublets at 8.06 and 6.63~ppm ($J=9.0$~Hz). As expected for a mixture of this simplicity, \nimis{} demixed it perfectly, assigning every peak to its correct component.

The second example is a more demanding binary mixture of 4-methylbenzaldehyde and the imine (\emph{E})-N-(4-bromobenzyl)-1-(4-methylphenyl)methanimine (Figure~\ref{fig:case2}) in which one component is a substructure of the other, so that the two molecules give very similar signals at close chemical shifts. The signals remain resolved and do not overlap, but the spectrum is noisier than that of the first example, as seen in the aromatic region and in the integral values. \nimis{} correctly assigned nine of the ten peaks. The single error is the methyl singlet at 2.46~ppm, which was assigned to the wrong component. This signal has no J-coupling, so this cue could not link it to its molecule, and it lies only 0.05~ppm from the methyl signal of the other component at 2.41~ppm. \nimis{} can therefore separate structurally related components in a noisy spectrum of moderate complexity without signal overlap.

The third example is a ternary mixture of 4-bromobenzaldehyde, 3,5-dimethoxybenzylamine and the imine formed by their condensation (Figure~\ref{fig:case3}). It is considerably more demanding than the binary examples, because the imine contains the structural fragments of both starting materials, so that many of its peaks overlap with benzaldehyde and amine peaks. Of the fourteen signals expected, the automatic annotation detected nine. The aldehyde proton of 4-bromobenzaldehyde at 9.98~ppm, which the software classified as an impurity and ignored, was added manually, and the resulting ten peaks were passed to \nimis. Because each input peak is assigned to a single component, a composite signal with contributions from more than one molecule cannot be assigned correctly in full, and a perfect demixing is therefore not expected. The groupings returned by \nimis{} (Table~S5 of the Supporting Information) are nevertheless chemically meaningful. The two most characteristic signals of the imine, the CH=N proton at 8.28~ppm and the neighbouring methylene at 4.76~ppm, both of which lie next to the imine bond, were assigned to the same component. The aldehyde proton was separated into a component of its own, and the methoxy resonance at 3.81~ppm was grouped with the dimethoxyphenyl aromatic signal at 6.48~ppm, as expected for the 3,5-dimethoxybenzyl fragment shared by the amine and the imine. Two aromatic signals were assigned to incorrect molecules; for these signals, the assignment appears to follow from the integral values alone. This example shows that \nimis{} can recover meaningful groupings from a spectrum with substantial overlap but cannot resolve the composite signals themselves, which is the limitation discussed above.

These results demonstrate the feasibility of combining automated spectral annotation with \nimis{} to recover component-specific signal groups from experimental mixture spectra while retaining the measured integral values. In the two binary examples no signals from different components overlapped, although several resonances were closely spaced and explicit multiplicity assignments and J-coupling constants were unavailable for some signals after automatic annotation. \nimis{} nevertheless grouped nearly all peaks correctly, which suggests that it can tolerate incomplete coupling information in an automated workflow; the single misassigned peak, a methyl singlet without J-coupling, shows where this tolerance reaches its limit. The ternary mixture, in contrast, contains extensive signal overlap, which \nimis{} cannot resolve, although it still recovers meaningful groupings. The small number and limited complexity of these examples provide only preliminary evidence of practical applicability and do not allow a comprehensive assessment of generalisation to experimental mixtures. They nevertheless support further development and evaluation of the framework on more complex mixtures, including cases in which overlapping resonances complicate both signal annotation and component separation.

\section{Conclusions}

This work introduces \nimis, a structure-agnostic framework for demixing 1D $^1$H-NMR spectra of chemical mixtures with chemically informed slot-attention clustering. From processed spectral data alone, without prior structural or system knowledge, the model groups peaks that originate from the same mixture component. Introducing NMR domain knowledge into the attention mechanism through a J-coupling matching matrix improves peak-to-component attribution, and \nimis{} surpasses both heuristic baselines and a conventional Transformer. The model reaches an assignment accuracy of 82.28\% under noise-free conditions and 77.34\% at 10\% integral noise, with a mean \nimis{} score of 0.83 for correct assignments. Performance is retained on a scaffold-disjoint test set, and accuracy degrades gradually under systematic errors in the input integrals. Experimental mixtures of moderate complexity are demixed largely correctly, whereas a ternary mixture with extensive signal overlap shows that \nimis, which assigns each peak to a single component, cannot resolve composite signals. \nimis{} is therefore an incomplete solution to mixture analysis, but it provides a foundation for developing complete methods for the direct computational analysis of mixtures, including crude reaction mixtures. Code and the trained model weights are available on GitHub at https://github.com/annamarijasumrova/NIMIS. The training, validation and test sets are available on Zenodo (DOI: 10.5281/zenodo.23961232).

\section*{Acknowledgements}

This work was supported by the University of Cambridge Harding Distinguished Postgraduate Scholars Programme.

The authors acknowledge that the experimental data used in this work were collected during the first author's Master's project at Imperial College London. The authors thank Benjamin J.\ Deadman and Christopher S.\ Roberts for scientific discussions and technical support. The authors additionally thank Professor Mimi (King Kuok) Hii for guidance and support during experimental data generation and subsequent data reuse in this study.

The authors also acknowledge access to the ROAR research facility and NMR infrastructure. The ROAR facility was funded through EPSRC Strategic Equipment Grants EP/R008825/1 and EP/V029037/1.

\end{document}


\maketitle

\tableofcontents
\bigskip

\section{Model architecture}
\label{sec:arch}

\nimis{} maps an unordered set of $P$ processed $^1$H-NMR peaks to a soft assignment of each peak to one of $K$ latent mixture components. It has three stages (main text, Fig.~1A): (i) a per-peak scalar embedding, (ii) a Transformer encoder whose attention logits are biased by a pairwise J-coupling matching matrix, and (iii) an iterative slot-attention module that clusters the contextualised peaks. The encoder uses no positional encoding in the sequence sense: peaks are an unordered set and chemical shift enters only through the embedding of Section~\ref{sec:embed}.

\subsection{Notation}
Let $B$ be the batch size, $P$ the number of peaks in a mixture (padded to $P_{\max}$ in a batch, with a boolean peak mask $m\in\{0,1\}^{P}$), $D=128$ the model width, $H=8$ the number of attention heads, $d_h=64$ the per-head width, $L=8$ the encoder depth, and $K=5$ the number of slots. Peak $i$ has chemical shift $\delta_i$ (ppm), integral $I_i$, and up to three J-couplings $\mathbf{J}_i\in\mathbb{R}^{3}$ (Hz) with binary validity mask $\mathbf{u}_i\in\{0,1\}^3$ (zero-padded, sorted descending).

\subsection{Peak embedding}
\label{sec:embed}
Each peak is represented by two scalars, $\delta_i$ and $I_i$. The shift is expanded with $n_f=16$ sine/cosine Fourier features,
\begin{equation}
\phi(\delta)=\big[\sin(\omega_1\delta),\dots,\sin(\omega_{16}\delta),\ \cos(\omega_1\delta),\dots,\cos(\omega_{16}\delta)\big],\qquad \omega_k=\frac{2\pi}{T_k},
\end{equation}
with periods $T_k$ geometrically spaced from $16$ ppm to $0.01$ ppm. The integral is compressed as $\log(1+\max(I_i,0))$, which keeps small (low-abundance) integrals approximately linear while compressing large ones. The resulting $2n_f+1=33$ features are mapped to $D$ dimensions by a three-layer MLP ($33\rightarrow64\rightarrow64\rightarrow128$, ReLU after the first two layers, no activation on the output). Dropout ($p=0.1$) is applied to the embedding.

\subsection{Pairwise bias matrices}
\label{sec:bias}
Two symmetric $P\times P$ matrices are computed from the peak list and fed to every attention layer. Neither is learned.

\textbf{J-coupling matching matrix.}
\begin{equation}
\CJ[i,j]=\mathbb{1}\big[\exists\,a,b:\ u_{ia}u_{jb}=1,\ |J_{ia}-J_{jb}|\le\tau\big]\cdot h_i h_j,
\end{equation}
where $\tau=0.1$~Hz, $h_i=\mathbb{1}[\sum_a u_{ia}>0]$ indicates that peak $i$ carries at least one coupling, and the diagonal is $1$ for all peaks. Peaks without reported couplings (including all singlets) therefore have a zero off-diagonal row and a self-entry of $1$.

\textbf{Integral-ratio matrix.} $\CI[i,j]=\min(I_i/I_j,\,I_j/I_i)\in[0,1]$, with $0$ if either integral is zero and $1$ on the diagonal. $\CI$ is invariant to a global rescaling of the integrals (Section~\ref{sec:gui}). It is used only in the ablations of the main text (Table~1); the released model does not use it.

\subsection{Attention with J-coupling bias}
\label{sec:attn}
Each encoder layer is a pre-LayerNorm Transformer block. For layer $\ell$ with input $X\in\mathbb{R}^{P\times D}$, queries, keys and values are computed from a single bias-free projection of $\mathrm{LN}(X)$, $[Q,K,V]=\mathrm{LN}(X)W_{qkv}$, with $W_{qkv}\in\mathbb{R}^{D\times 3Hd_h}$, split into $H$ heads. Per head,
\begin{equation}
A^{(\ell)}=\mathrm{softmax}\!\Big(\tfrac{QK^\top}{\sqrt{d_h}}+w_J^{(\ell)}\,\CJ+w_I^{(\ell)}\,\CI+M\Big)V,
\end{equation}
where $M$ is $-10^{9}$ at padded key positions and $0$ elsewhere. $w_J^{(\ell)}\in\mathbb{R}$ is a learnable scalar per layer, shared by all heads and initialised to $1$ (the same bias is added to every head; a per-head variant was implemented but not used). For the released configuration the integral-ratio term is disabled ($\CI$ is not passed, so $w_I^{(\ell)}$ stays at its initial value and has no effect). Head outputs are concatenated and projected by a bias-free $W_o\in\mathbb{R}^{Hd_h\times D}$ followed by dropout ($p=0.1$). The block is
\begin{align}
X' &= X + \mathrm{Dropout}\big(W_o\,\mathrm{concat}_h(A^{(\ell)}_h)\big),\\
X'' &= X' + W_2\,\mathrm{GELU}\big(W_1\,\mathrm{LN}(X')\big),
\end{align}
with $W_1\in\mathbb{R}^{D\times 256}$, $W_2\in\mathbb{R}^{256\times D}$ (both with bias), i.e.\ an MLP expansion ratio of 2. The encoder stacks $L=8$ such blocks and returns the final $X''$ (no final norm). If $\CJ\equiv0$ the block reduces exactly to a standard pre-LN Transformer layer, which is why the vanilla Transformer baseline is obtained by simply omitting $\CJ$. When no peak carries a coupling, $\CJ$ is the identity matrix rather than zero, so the block differs from the standard layer only by the constant self-attention term $w_J^{(\ell)}$ added to the logit of each peak attending to itself.

Table~\ref{tab:wj} lists the learned $w_J^{(\ell)}$ of the released checkpoint. They differ in sign across depth (strongly positive in layers 1 and 3, negative in the middle layers), so the bias is used as a layer-dependent signal and not as a uniform attraction between J-matched peaks.

\begin{table}[h]
\centering
\caption{Learned J-coupling bias weight $w_J^{(\ell)}$ per encoder layer in the released checkpoint.}
\label{tab:wj}
\begin{tabular}{lcccccccc}
\toprule
Layer $\ell$ & 1 & 2 & 3 & 4 & 5 & 6 & 7 & 8\\
\midrule
$w_J^{(\ell)}$ & 2.86 & $-0.84$ & 2.24 & 0.19 & $-0.08$ & $-0.95$ & $-0.76$ & 1.51\\
\bottomrule
\end{tabular}
\end{table}

\subsection{Slot attention}
\label{sec:slot}
The contextualised peak features $X\in\mathbb{R}^{P\times D}$ are clustered into $K=5$ slots by slot attention with $T=5$ iterations. Slots are initialised by sampling $s_k\sim\mathcal{N}(\mu,\mathrm{diag}(\mathrm{softplus}(\sigma)^2))$ with learned $\mu,\sigma\in\mathbb{R}^{D}$ shared across slots (so slot identity is exchangeable and the output is defined only up to permutation). Inputs are normalised by LayerNorm and projected to keys and values, $k=W_k\mathrm{LN}(X)$, $v=W_v\mathrm{LN}(X)$ ($D\times D$, bias-free). At iteration $t$:
\begin{align}
q_k &= W_q\,\mathrm{LN}(s_k),\\
a_{kp} &= \mathrm{softmax}_{k}\!\Big(\tfrac{q_k\cdot k_p}{\sqrt{D}}\Big)\quad\text{(over slots, not peaks)},\\
u_k &= \sum_{p} a_{kp}\,v_p,\\
s_k &\leftarrow \mathrm{GRU}(u_k,\ s_k),\qquad s_k\leftarrow s_k+\mathrm{MLP}(\mathrm{LN}(s_k)).
\end{align}
The GRU is a standard \texttt{GRUCell} ($D\rightarrow D$) and the residual MLP is $D\rightarrow 2D\rightarrow D$ with GELU. The softmax over slots makes slots compete for each peak. The update $u_k$ is an un-normalised attention-weighted sum rather than a weighted mean over peaks. Padded peaks are masked out of the logits, and slots beyond the number of components $n$ of the mixture are masked out of the softmax with $-10^{9}$ via a component mask. Hence the number of mixture components $n\in\{2,\dots,5\}$ is an \emph{input} to the model at both training and inference time, and only $K_{\max}=5$ is a fixed capacity.

\subsection{Output and NIMIS score}
The output is the final-iteration attention matrix $a\in\mathbb{R}^{K\times P}$. Peak $p$ is assigned to $\hat k_p=\arg\max_k a_{kp}$, and its \emph{NIMIS score} is $a_{\hat k_p,p}\in(0,1]$. Because the softmax runs over the $n$ active slots, the score is lower-bounded by $1/n$. Slot indices are arbitrary; to evaluate against ground truth they are aligned to the true components with the Hungarian algorithm (Section~\ref{sec:loss}).

\subsection{Size summary}

\begin{table}[h]
\centering
\caption{Hyperparameters and parameter counts of the released \nimis{} model.}
\label{tab:arch}
\begin{tabular}{lr}
\toprule
\textbf{Component} & \textbf{Setting}\\
\midrule
Model width $D$ & 128\\
Peak input features (Fourier $2\times16$ (peak position) + log-integral) & 33\\
Embedding MLP (hidden widths) & $33\rightarrow64\rightarrow64\rightarrow128$\\
Fourier periods & 16 to 0.01 ppm, 16 geometric steps\\
Encoder depth $L$ & 8\\
Attention heads $H$ / head width $d_h$ & 8 / 64 (inner width 512)\\
Feed-forward width / activation & 256 / GELU\\
Normalisation & pre-LayerNorm\\
Dropout (embedding, attention output) & 0.1\\
Attention bias & $w_J^{(\ell)}\CJ$, one scalar per layer, $\tau=0.1$ Hz\\
Max.\ J-couplings per peak & 3\\
Slots $K$ / slot-attention iterations $T$ & 5 / 5\\
Slot update & GRUCell + residual MLP ($D\rightarrow2D\rightarrow D$, GELU)\\
Slot initialisation & learned Gaussian $(\mu,\sigma)$, sampled\\
\midrule
Parameters: embedding & 14{,}656\\
Parameters: encoder (8 layers $\times$ 328{,}578) & 2{,}628{,}624\\
Parameters: slot attention & 215{,}168\\
\textbf{Total trainable parameters} & \textbf{2{,}858{,}448}\\
\bottomrule
\end{tabular}
\end{table}
The count includes the eight unused $w_I^{(\ell)}$ scalars. Each encoder layer holds 196{,}608 (QKV) + 65{,}536 ($W_o$) + 65{,}920 (feed-forward) + 514 (two LayerNorms and two bias scalars) parameters. The model is relatively small (about 2.9\,M parameters) and runs on CPU in the GUI (Section~\ref{sec:gui}).

\section{Training objective}
\label{sec:loss}
Slot order is arbitrary, so the loss is made permutation invariant by matching slots to ground-truth components for every mixture. For a mixture with $n$ true components and true labels $y_p\in\{1,\dots,n\}$, define the cost of assigning slot $k$ to component $c$ as the mean negative log-probability that the slot's attention gives to the peaks of $c$,
\begin{equation}
\mathrm{cost}[k,c]=-\frac{1}{|\mathcal{P}_c|}\sum_{p\in\mathcal{P}_c}\log\big(a_{kp}+10^{-9}\big),\qquad \mathcal{P}_c=\{p:y_p=c\}.
\end{equation}
The optimal one-to-one matching $\pi$ is found with the Hungarian algorithm (\texttt{scipy.optimize.linear\_sum\_assignment}) on the $K\times n$ cost matrix, treated as a constant. The loss is the cross-entropy of the matched slot, averaged over all valid peaks in the batch,
\begin{equation}
\mathcal{L}=-\frac{1}{\sum_b P_b}\sum_{b}\sum_{p\le P_b}\log\max\big(a^{(b)}_{\pi_b(y_p),p},\,10^{-9}\big).
\end{equation}
Gradients flow through $a$ only. The same matching is used at test time to compute assignment accuracy (fraction of peaks whose matched slot equals the true component) and confidence statistics.

\section{Training and evaluation protocol}
\label{sec:train}
\begin{table}[h]
\centering
\caption{Optimisation settings.}
\label{tab:opt}
\begin{tabular}{ll}
\toprule
Optimiser & Adam, initial learning rate $1\times10^{-4}$\\
LR schedule & ReduceLROnPlateau on epoch training loss (factor 0.5, patience 8, min $5\times10^{-6}$)\\
Gradient clipping & global norm 1.0\\
Batch size & 128 per GPU (data-parallel via PyTorch DDP when several GPUs are used)\\
Epochs & 300 (validation every 5 epochs, best-validation-loss checkpoint kept)\\
Integral augmentation & $I\leftarrow I\,(1+\epsilon)$, $\epsilon\sim\mathcal{N}(0,0.1^2)$ per peak, redrawn every batch\\
Test-time integral noise & 0, 5, 10, 15\,\%\\
Training data & 158{,}005 mixtures of 2--5 components (about 39.5\,k per size)\\
Framework & PyTorch 2.8.0\\
\bottomrule
\end{tabular}
\end{table}

Integral noise is applied on the fly in the batch-collation step, and $\CI$ (when used) is computed from the \emph{noisy} integrals so that both inputs are consistent. $\CJ$ is unaffected by integral noise. Mixtures in a batch are padded to the longest peak list and masked. Dataset construction, splits, and evaluation sets are described in the main text (Methods).

\section{Detailed model performance}
\label{sec:perf}
Table~\ref{tab:perf} breaks down the peak-to-component assignment accuracy of the released \nimis{} model by the number of mixture components and by the integral noise added at test time, on the molecule-disjoint and the (deduplicated) scaffold-disjoint test sets. Accuracy is the fraction of peaks whose Hungarian-matched slot equals the true component (Section~\ref{sec:loss}), and the NIMIS score columns give the mean score of correctly and incorrectly assigned peaks. Accuracy falls as mixtures contain more components, from 91.5\,\% for two components to 69.0\,\% for five at 10\,\% noise on the molecule-disjoint set, and it falls with integral noise, from 82.28\,\% without noise to 72.44\,\% at 15\,\% noise. The mean score of correct assignments stays at about 0.83 at every noise level, whereas that of incorrect assignments rises from 0.57 to 0.62, so the confidence separation narrows modestly with noise. The scaffold-disjoint accuracies are close to the molecule-disjoint ones at every noise level and mixture size; the largest differences occur at 15\,\% noise for four components (71.10\,\% versus 70.63\,\%) and five components (64.19\,\% versus 63.72\,\%).

\begin{table}[h]
\centering
\small
\setlength{\tabcolsep}{4.5pt}
\caption{Peak-to-component assignment accuracy (\%) of the released model by number of mixture components and added integral noise, with the mean NIMIS score of correctly and incorrectly assigned peaks.}
\label{tab:perf}
\begin{tabular}{rccccccc}
\toprule
Noise (\%) & 2 & 3 & 4 & 5 & All & \multicolumn{2}{c}{Mean NIMIS score}\\
\cmidrule(lr){7-8}
 & comp. & comp. & comp. & comp. & & correct & incorrect\\
\midrule
\multicolumn{8}{l}{\textit{Molecule-disjoint test set (16{,}000 mixtures, 388{,}646 peaks)}}\\
0  & 93.75 & 87.62 & 81.71 & 74.89 & 82.28 & 0.832 & 0.571\\
5  & 93.37 & 86.62 & 80.50 & 73.48 & 81.16 & 0.830 & 0.573\\
10 & 91.47 & 83.28 & 76.12 & 69.04 & 77.34 & 0.828 & 0.590\\
15 & 88.42 & 78.63 & 70.63 & 63.72 & 72.44 & 0.828 & 0.618\\
Peaks & 55{,}775 & 83{,}429 & 110{,}919 & 138{,}523 & 388{,}646 & &\\
\midrule
\multicolumn{8}{l}{\textit{Scaffold-disjoint test set (15{,}991 mixtures, 388{,}688 peaks)}}\\
0  & 93.79 & 87.77 & 82.04 & 75.28 & 82.54 & 0.831 & 0.569\\
5  & 93.36 & 86.96 & 80.87 & 73.80 & 81.44 & 0.828 & 0.570\\
10 & 91.57 & 83.54 & 76.54 & 69.49 & 77.68 & 0.826 & 0.587\\
15 & 88.29 & 79.02 & 71.10 & 64.19 & 72.80 & 0.825 & 0.615\\
Peaks & 55{,}785 & 83{,}195 & 110{,}983 & 138{,}725 & 388{,}688 & &\\
\bottomrule
\end{tabular}
\end{table}

\section{Experimental ternary mixture: assignment details}
\label{sec:ternary}
Table~\ref{tab:ternary} lists the peak-to-component assignment produced by the released \nimis{} model for the experimental ternary mixture of 4-bromobenzaldehyde, 3,5-dimethoxybenzylamine and their condensation product (the imine) shown in the main text. Peaks were obtained with the automatic annotation tool of MestReNova, and the unrounded integral values displayed under each peak were passed to \nimis{} unchanged (Section~\ref{sec:inputs}). The automatic analysis detected nine of the 14 expected signals; the aldehyde proton at 9.98\,ppm, which the software classified as an impurity and ignored, was added manually, so ten peaks were supplied to the model. Component numbers are arbitrary labels of the predicted groups, and the NIMIS score is the confidence of the assignment of each peak. Because the imine contains the structural fragments of both other components, several signals (for example the methoxy and the dimethoxyphenyl resonances) contain contributions from more than one molecule, and a peak can be assigned to only one component.

\begin{table}[h]
\centering
\small
\setlength{\tabcolsep}{8pt}
\caption{Peak-to-component assignment of the experimental ternary mixture by the released \nimis{} model. Integrals are the unrounded values determined by the automatic annotation and displayed under each peak (the aldehyde peak at 9.98\,ppm was added manually).}
\label{tab:ternary}
\begin{tabular}{clcc}
\toprule
Component & Peak position (ppm) & Integral & NIMIS score\\
\midrule
1 & 4.76 & 2.08  & 0.6296\\
  & 7.12 & 0.79  & 0.5711\\
  & 7.19 & 0.99  & 0.6060\\
  & 7.55 & 1.91  & 0.6058\\
  & 7.64 & 2.04  & 0.5419\\
  & 8.28 & 0.87  & 0.5819\\
\midrule
2 & 3.81 & 15.36 & 0.9457\\
  & 6.48 & 4.30  & 0.7794\\
  & 7.73 & 3.07  & 0.5438\\
\midrule
3 & 9.98 & 0.60  & 0.7580\\
\bottomrule
\end{tabular}
\end{table}

\section{Robustness to missing J-coupling information}
\label{sec:jdrop}
Experimental peak lists often contain few or no extractable coupling constants. To quantify how \nimis{} behaves when J-coupling information is lost, the released model was re-evaluated without retraining while a random fraction of the J-coupling matches was removed at test time. For each mixture, the given fraction of the off-diagonal entries equal to 1 in $\CJ$ was set to 0, symmetrically so that $\CJ$ stayed symmetric, and the diagonal was left unchanged. Removing 0\,\% of the matches is the unmodified input, and removing 100\,\% leaves a matrix that contains only its diagonal, i.e.\ a spectrum without any usable coupling information. Evaluation used the molecule-disjoint test set with 10\,\% integral noise, and the results in Table~\ref{tab:jdrop} are means and standard deviations over five evaluation seeds, which resample both the integral noise and the removal mask.

Accuracy decreases gradually and without abrupt failure as matches are removed: from 77.33\,\% with all matches to 77.00\,\% with 25\,\% removed, 76.62\,\% with 50\,\% removed and 75.74\,\% with all matches removed. The decrease is similar for every number of components; for five-component mixtures, accuracy falls from 68.98\,\% to 67.10\,\%. Even when no J-coupling matches remain, accuracy (75.74\,\%) stays above that of the vanilla Transformer (74.61\,\% at 10\,\% noise, Table~1 of the main text), although the model was never exposed to J-free input during training. These results indicate that \nimis{} degrades gradually towards the performance of an unbiased Transformer, and does not fail, when coupling information is incomplete.

\begin{table}[h]
\centering
\footnotesize
\setlength{\tabcolsep}{3pt}
\caption{Peak-to-component assignment accuracy (\%) of the released model on the molecule-disjoint test set (10\,\% integral noise) when a fraction of the J-coupling matches is removed at test time, overall and by number of mixture components. The peaks-matched column is the fraction of peaks with at least one remaining match. Values are means $\pm$ standard deviations over five evaluation seeds.}
\label{tab:jdrop}
\begin{tabular}{ccccccc}
\toprule
Matches removed (\%) & Peaks matched (\%) & All & 2 comp. & 3 comp. & 4 comp. & 5 comp.\\
\midrule
0 & 26.68 & 77.33 $\pm$ 0.03 & 91.42 $\pm$ 0.10 & 83.25 $\pm$ 0.04 & 76.22 $\pm$ 0.11 & 68.98 $\pm$ 0.04\\
10 & 25.06 & 77.21 $\pm$ 0.06 & 91.36 $\pm$ 0.09 & 83.14 $\pm$ 0.09 & 76.06 $\pm$ 0.13 & 68.87 $\pm$ 0.06\\
25 & 22.41 & 77.00 $\pm$ 0.04 & 91.22 $\pm$ 0.09 & 82.99 $\pm$ 0.06 & 75.85 $\pm$ 0.10 & 68.58 $\pm$ 0.07\\
50 & 17.08 & 76.62 $\pm$ 0.03 & 91.06 $\pm$ 0.04 & 82.57 $\pm$ 0.05 & 75.41 $\pm$ 0.13 & 68.18 $\pm$ 0.07\\
75 & 9.99 & 76.21 $\pm$ 0.05 & 90.89 $\pm$ 0.08 & 82.24 $\pm$ 0.04 & 74.93 $\pm$ 0.15 & 67.68 $\pm$ 0.09\\
100 & 0.00 & 75.74 $\pm$ 0.04 & 90.67 $\pm$ 0.08 & 81.85 $\pm$ 0.04 & 74.43 $\pm$ 0.08 & 67.10 $\pm$ 0.07\\
\bottomrule
\end{tabular}
\end{table}

\section{Experimental mixtures: \nimis{} input peak lists}
\label{sec:inputs}
Table~\ref{tab:inputs} lists the complete inputs supplied to \nimis{} for the three experimental mixtures in the main text (Figures~7--9). All spectra were recorded at 400\,MHz in CDCl$_3$ and annotated with the automatic analysis tool of MestReNova. Peak positions are the peak labels of the automatic analysis, and multiplicities and J-coupling constants are taken from the automatically generated multiplet report shown above each spectrum. That report rounds the proton count of each signal to a whole number (column ``Report''). Because these rounded values discard the relative abundance of the components, the unrounded integrals determined by the same analysis and displayed under each peak in the spectra (column ``Integral used'') were supplied to \nimis{} instead. Signals reported as multiplets (m) without resolved coupling constants were supplied without J-values. For the ternary mixture, the aldehyde signal at 9.98\,ppm, which the software classified as an impurity, was added manually with the integral displayed under the peak; all other values were used without correction.

Each mixture can be entered in the \nimis{} interface (Section~\ref{sec:gui}) as one row per peak with the values in Table~\ref{tab:inputs}, or as an SI-style string in which the proton counts are replaced by the integrals used, e.g.\ for the first mixture:

\begin{quote}\small\ttfamily
9.89 (s, 1.00H), 8.06 (d, J = 9.0 Hz, 0.69H), 7.85 (d, J = 8.8 Hz, 2.28H), 7.02 (d, J = 8.8 Hz, 2.27H), 6.63 (d, J = 9.0 Hz, 0.67H), 3.90 (s, 3.68H)
\end{quote}

\begin{table}[h]
\centering
\small
\setlength{\tabcolsep}{7pt}
\caption{\nimis{} input peak lists for the experimental mixtures of main-text Figures~7--9 (400\,MHz, CDCl$_3$). ``Report'' is the rounded proton count in the automatically generated multiplet report; ``Integral used'' is the unrounded integral displayed under the peak and supplied to \nimis.}
\label{tab:inputs}
\begin{tabular}{ccccc}
\toprule
Peak position (ppm) & Multiplicity & J (Hz) & Report (H) & Integral used\\
\midrule
\multicolumn{5}{l}{\textit{Binary mixture: 4-methoxybenzaldehyde and 4-nitroaniline (main-text Figure 7)}}\\
9.89 & s & -- & 1 & 1.00\\
8.06 & d & 9.0 & 1 & 0.69\\
7.85 & d & 8.8 & 2 & 2.28\\
7.02 & d & 8.8 & 2 & 2.27\\
6.63 & d & 9.0 & 1 & 0.67\\
3.90 & s & -- & 4 & 3.68\\
\midrule
\multicolumn{5}{l}{\textit{Binary mixture: 4-methylbenzaldehyde and (E)-N-(4-bromobenzyl)-1-(4-methylphenyl)methanimine (main-text Figure 8)}}\\
9.98 & s & -- & 1 & 0.90\\
8.37 & d & 1.5 & 2 & 2.28\\
7.80 & d & 8.1 & 2 & 1.81\\
7.70 & d & 7.9 & 3 & 3.25\\
7.48 & m & -- & 4 & 4.06\\
7.35 & d & 7.8 & 2 & 1.84\\
7.24 & dd & 8.2, 6.3 & 8 & 7.65\\
4.76 & s & -- & 3 & 3.50\\
2.46 & s & -- & 3 & 3.28\\
2.41 & s & -- & 5 & 4.53\\
\midrule
\multicolumn{5}{l}{\textit{Ternary mixture: 4-bromobenzaldehyde, 3,5-dimethoxybenzylamine and imine (main-text Figure 9)}}\\
9.98$^{a}$ & s & -- & -- & 0.60\\
8.28 & s & -- & 1 & 0.87\\
7.73 & m & -- & 3 & 3.07\\
7.64 & m & -- & 2 & 2.04\\
7.55 & m & -- & 2 & 1.91\\
7.19 & m & -- & 1 & 0.99\\
7.12 & d & 8.1 & 1 & 0.79\\
6.48 & m & -- & 4 & 4.30\\
4.76 & d & 1.5 & 2 & 2.08\\
3.81 & m & -- & 15 & 15.36\\
\bottomrule
\multicolumn{5}{l}{\footnotesize $^{a}$Classified as an impurity by the automatic analysis and added manually.}
\end{tabular}
\end{table}
\FloatBarrier

\section{Graphical user interface (GUI)}
\label{sec:gui}
The released checkpoint is wrapped in a Streamlit application (\texttt{gui/nimis\_gui.py}, run with \texttt{streamlit run gui/nimis\_gui.py}) so that  \nimis{} can be used directly on processed peak. The model runs on CPU.

\subsection{Workflow}
\begin{enumerate}
\item \textbf{Input.} The user either pastes an SI-style $^1$H-NMR string, e.g.\ \texttt{7.26 (d, J = 8.1 Hz, 2H), 7.12--7.05 (m, 3H), 2.31 (s, 3H)}, or enters one row per peak in a table (position in ppm, integral, and optional $J_1$--$J_3$ in Hz). Position and integral are required, J-couplings are optional.
\item \textbf{Parsing and review.} An SI string is parsed with regular expressions into position, integral (the number preceding \texttt{H}; 1.0 if absent), and up to three J-values. The parsed peaks are shown in an editable table so that the user can correct mis-parsed values before running. Rows with a missing or non-numeric position/integral are skipped, and the skipped row numbers are reported.
\item \textbf{Number of components.} The user selects the number of mixture components $n$. This sets the component mask of Section~\ref{sec:slot}.
\item \textbf{Pre-processing.} Peaks are sorted by shift, $\CJ$ is built with $\tau=0.1$~Hz exactly as in training, and the integrals are rescaled as described below.
\item \textbf{Inference.} A single forward pass is run with a fixed random seed (slot initialisation is stochastic), and the output is $\hat k_p$ and the NIMIS score for each peak.
\item \textbf{Output.} Peaks are listed per predicted component (component numbering is arbitrary) with their position, \emph{original} user-supplied integral, and NIMIS score.
\end{enumerate}

\subsection{Agnosticism to the integral scale}
Experimental integrals are reported in arbitrary units: an integral may be normalised to a reference proton ($=1$), to the total proton count, or to the largest peak, and the same mixture can therefore be entered at very different numerical scales. The training integrals, in contrast, carry the mixture proportions: a peak's integral is its proton count multiplied by its component's proportion, and the proportions sum to one. Peak integrals in the training set therefore span roughly $0.02$--$3$ (1st--99th percentile $0.017$--$3.2$), with a median of $0.35$. Because the embedding of Section~\ref{sec:embed} is not scale invariant (it contains $\log(1+I)$), feeding integrals at an arbitrary scale would present the network with values from outside its training distribution even when all ratios are correct.

The GUI therefore removes the dependence on the input scale by a single global rescaling,
\begin{equation}
\tilde I_i \;=\; I_i\cdot\frac{I_\star}{\operatorname{median}_j(I_j)},\qquad I_\star=0.356,
\label{eq:remap}
\end{equation}
where $I_\star$ is the median, over training mixtures, of the per-mixture median peak integral (we measure $0.355$; the pooled per-peak median is $0.348$). The rescaled values $\tilde I$ are used only in the embedding. Properties of this mapping:
\begin{itemize}
\item \textbf{Scale invariance.} For any $s>0$, $I\mapsto sI$ gives the same $\tilde I$, because the median scales with $I$. Entering a spectrum with integrals $(1,2,3)$ or $(10,20,30)$ gives identical network inputs.
\item \textbf{Ratios are preserved.} A global rescaling leaves all ratios $I_i/I_j$ unchanged, so no information about relative proton counts or abundances is lost. For this reason $\CI$ is computed from the \emph{original} integrals and is independent of the remapping (and is anyway not used by the released model).
\item \textbf{Anchoring.} After rescaling, the median input integral equals the training median, so a typical peak lies in the central part of the range seen during training. The spread of integrals around the median is determined by the user's data and is not altered.
\item \textbf{Reported values.} The output table displays the user's original integrals, not $\tilde I$.
\end{itemize}

\subsection{Limitations of the interface}
The model is trained with 10\,\% multiplicative integral noise, so it tolerates moderate integration error (main text, Table~1), but the median anchor in Eq.~\eqref{eq:remap} assumes that the median peak of the input is representative of the training distribution. Integrals dominated by a few very large peaks (e.g.\ an unsuppressed residual solvent or water signal included in the peak list) shift the median and therefore all rescaled values, and such peaks should be removed before running. The interface also requires $n$ from the user, takes J-values only from the first three couplings reported, and parses multiplets by their first listed shift. Automatic annotation tools can produce the same peak-list format.